\documentclass[fleqn,11pt]{wlscirep}
\usepackage[utf8]{inputenc}
\usepackage[T1]{fontenc}

\usepackage[utf8]{inputenc} 
\usepackage[T1]{fontenc}    
\usepackage{url}            
\usepackage{booktabs}       
\usepackage{amsfonts}       
\usepackage{nicefrac}       
\usepackage{microtype}      
\usepackage{xcolor}         
\usepackage{amsmath}
\usepackage{siunitx}
\usepackage{graphicx}
\usepackage{makecell}
\usepackage[colorlinks=true]{hyperref}
\usepackage{bm}
\usepackage[
    natbib=true,
    style=nature,
    sorting=none,
    hyperref=true
]{biblatex}
\DeclareFieldFormat{urldate}{}
\AtEveryBibitem{\clearfield{urlyear}}
\AtEveryBibitem{
  \clearfield{note}
  \clearfield{issn}
  \clearfield{language}
}
\usepackage{booktabs}
\usepackage{multirow}
\usepackage{bbm}

\usepackage{algorithm}
\usepackage{algpseudocode}
\usepackage{rotating}

\DeclareUnicodeCharacter{0301}{\'{e}}
\usepackage{subcaption}
\usepackage{caption}

\usepackage{geometry}

\usepackage{inconsolata}
\usepackage[skip=6pt plus1pt, indent=0pt]{parskip}
\newcommand{\beginsupplement}{%
        \setcounter{table}{0}
        \renewcommand{\thetable}{S\arabic{table}}%
        \setcounter{figure}{0}
        \renewcommand{\thefigure}{S\arabic{figure}}%
        \setcounter{section}{0}
        \renewcommand{\thesection}{\hspace{-6pt}}
        \setcounter{subsection}{0}
        \renewcommand{\thesubsection}{S.\arabic{subsection}}%
        \setcounter{algorithm}{0}
        \renewcommand{\thealgorithm}{S\arabic{algorithm}}
     }
\usepackage[capitalize,noabbrev,nameinlink]{cleveref}

\begin{document}

\title{NeuralPLexer3: Accurate Biomolecular Complex Structure Prediction with Flow Models}

\author[ ]{Zhuoran Qiao$^{*,\dagger}$, Feizhi Ding$^{*}$, Thomas Dresselhaus$^{*}$, Mia A. Rosenfeld$^{*}$, Xiaotian Han$^{*}$, Owen Howell, Aniketh Iyengar, Stephen Opalenski, Anders S. Christensen, Sai Krishna Sirumalla, Frederick R. Manby, Thomas F. Miller III, Matthew Welborn$^{*,\dagger}$}
\affil[ ]{Iambic Therapeutics, San Diego, CA 92121}
\affil[*]{Core Contributors.}
\affil[$\dagger$]{Correspondence to: zhuoran.qiao@iambic.ai; matt@iambic.ai. }

\begin{abstract}
Structure determination is essential to a mechanistic understanding of diseases and the development of novel therapeutics. Machine-learning-based structure prediction methods have made significant advancements by computationally predicting protein and bioassembly structures from sequences and molecular topology alone. Despite substantial progress in the field, challenges remain to deliver structure prediction models to real-world drug discovery. Here, we present NeuralPLexer3 -- a physics-inspired flow-based generative model that achieves state-of-the-art prediction accuracy on key biomolecular interaction types and improves training and sampling efficiency compared to its predecessors and alternative methodologies~\cite{abramson_accurate_2024,neuralplexer2}. Examined through newly developed benchmarking strategies, NeuralPLexer3 excels in vital areas that are crucial to structure-based drug design, such as physical validity and ligand-induced conformational changes. 
\end{abstract}

\flushbottom
\maketitle

\section*{Introduction}
For decades, predicting the 3D structures of proteins has been a transformative goal in structural biology and drug discovery. Experimental techniques like X-ray crystallography and cryo-electron microscopy have provided invaluable structural data. Still, these methods are resource-intensive and time-consuming, making it challenging to scale their application across the immense diversity of proteins and small molecules. Most drug development programs still rely on these experimental structures, leaving countless therapeutic opportunities and hypotheses unexplored. 

Breakthroughs in AI-driven structure prediction, notably with AlphaFold2 (AF2)~\cite{jumper_highly_2021}, brought the field closer to experimental accuracy.  Despite this progress, a significant challenge remained, that is, to accurately model interactions between proteins and different biomolecules, particularly small molecules, nucleic acids, and other proteins. Understanding and accurately modeling these interactions is essential for adapting structure prediction models into effective drug discovery workflows. 

NeuralPLexer (NP)~\citep{qiao2024state} was one of the first AI models to directly address these complex interactions by pioneering the use of AI for protein-ligand structures, with NeuralPLexer2 (NP2)~\cite{neuralplexer2} advancing further in prediction accuracy and covering all essential categories of biomolecules including protein complexes, nucleic acids, small molecules, and covalent modifications.  

AlphaFold3 (AF3)~\citep{abramson_accurate_2024} recently set a new benchmark in interaction modeling. While taking us another step further for drug discovery applications, this geometrical approach to interaction prediction still limits its usability. Drug discovery requires precise, atom-level details on how small molecules interact with specific protein atoms and how these interactions induce conformational changes. Rapid screening of large numbers of potential compounds is also necessary.  

Specifically, several technical challenges persist in state-of-the-art structure prediction models: 
\begin{itemize}
    \item \textbf{Unphysical hallucinations}: AF3 and related methods~\citep{abramson_accurate_2024,krishna2024generalized,lu2024dynamicbind} sometimes generate structures that are not physically plausible, such as ligands with incorrect chiral centers, unrealistic torsion angles, or misfolded chains for disordered regions.  
    \item \textbf{Computational cost}: The resource-intensive nature of current diffusion-based structure predictors can limit their scalability in large-scale studies, such as virtual screening. 
    \item \textbf{Performance assessment for drug discovery}: There is a need for thorough model evaluation in drug discovery contexts, particularly in predicting ligand-induced protein conformational changes and the recovery of physical interactions such as hydrogen bonds~\citep{errington_assessing_2024,mirdita_colabfold_2022}. 
\end{itemize}

Here we present NeuralPLexer3 (NP3), offering the following key contributions:  

\begin{itemize}
    \item NP3 improves protein-ligand binding structure prediction with greater accuracy than AF3 while retaining broad applicability across all categories of biomolecular interactions. 
    \item By combining physics-informed priors, fast samplers with hardware-aware optimizations, NP3 typically delivers a prediction within seconds of GPU time while preserving accuracy. 
    \item We introduce novel benchmarks to evaluate conformational predictions related to binding pocket/ligand interactions and their potential effects on protein conformation across a diverse dataset. For specific protein classes, such as kinases, NP3 provides downstream functional insights, such as predicting protein inactivation from ligand binding. 
\end{itemize}

We also present scaling studies for flow-based encoder-decoder structure prediction models and estimate the compute-optimal frontier, improving training efficiency relative to AF3 and related methods. This detailed breakdown of model performance offers insights about learning behaviors across various biomolecular modalities and identifies future areas for improvement.  

By supporting target validation, structural hypothesis generation, and atom-level interaction analysis, we seek to bridge critical gaps in AI-driven drug discovery, empowering researchers to navigate the complexities of therapeutic development with greater precision and speed.

\section*{Results}

\begin{figure}[tp]
    \centering
    \includegraphics[width=0.90\linewidth]{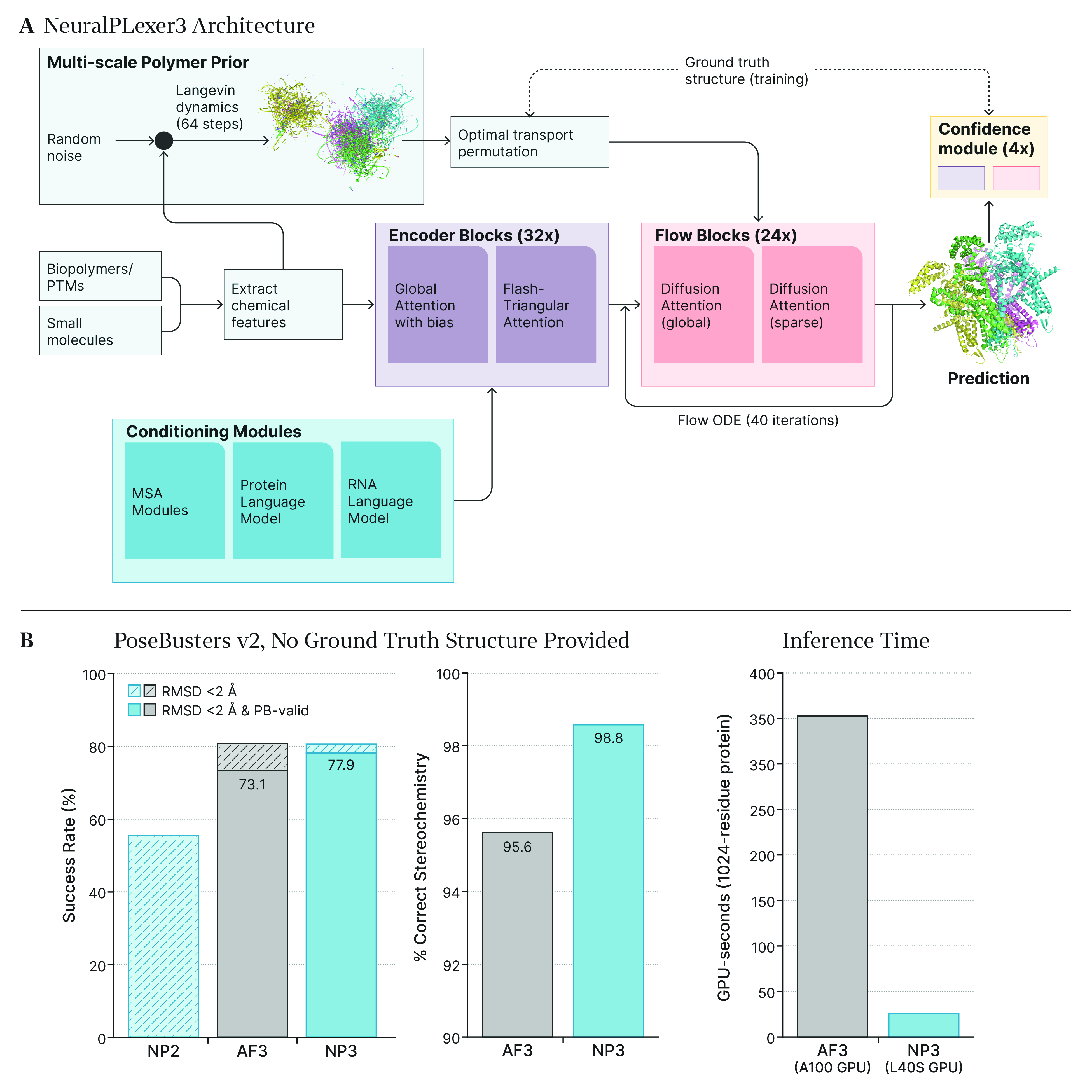}
    \caption{\textbf{NeuralPLexer3 (NP3) accurately predicts biomolecular structures with improved physical quality and prediction speed.} (\textbf{A}) Schematics of the NP3 system. To perform a prediction, NP3 uses molecular topology extracted from input biopolymer sequences and small molecule graphs as primitive inputs, with additional conditioning signals from sequence language models, multiple sequence alignments (MSAs), and templates. NP3 adopts a flow-matching framework that samples from an informative prior, which includes basic physical restraints on atom configurations, relaxed by Langevin dynamics using a globular polymer model with harmonic connectivity terms (\nameref{supp:si}, \cref{algo:a3:Globular}). During training, the prior samples are permuted to better align with the ground truth structure. (\textbf{B}) Performance of NP3 on the PoseBusters benchmark~\citep{buttenschoen_posebusters_2024}. Left panel: success rate for predicting the ligand-protein structures to within 2 \AA \space RMSD, with and without additionally requiring that the structures are physically reasonable (PB-valid). Center panel: percentage of predicted structures where ligand stereochemistry is correct. Right panel: timing for running a single inference on a 1024-residue protein. Comparisons are made to NeuralPLexer2 (NP2) and AlphaFold 3 (AF3).}
    \label{fig:np3_overview}
\end{figure}

NeuralPLexer3 (NP3) (\cref{fig:np3_overview}A) is a generative modeling framework for the \textit{de novo} prediction and sampling of generalized biomolecular complex structures composed of, but not limited to, proteins, nucleic acids, ligands, ions, and post-translational modifications (PTMs). The system is also designed to provide all-atom and pairwise confidence estimation of predicted complexes.

\subsection*{The NP3 main network}

Underpinning NP3 is a conditional flow-based generative model that samples the 3D coordinates of all heavy atoms of the complex from a model distribution~\citep{lipman_flow_2022,albergo_stochastic_2023}.

\textbf{Continuous normalizing flows}: The core model of NP3 is a flow-based generative model that utilizes continuous normalizing flows (CNFs)~\citep{lipman_flow_2022} and is trained without the use of simulations. CNFs transform simple probability distributions into complex ones through continuous-time dynamics. CNFs sample new data by integrating an ordinary differential equation (ODE) using initial conditions from the prior distribution: 
\begin{equation}
    x_t = x_0 + \int_0^t u_s(x_s) \, ds
\end{equation}

\textbf{Flow matching}: Flow matching enables efficient and stable training of CNFs by aligning the model's vector field with a target vector field derived from predefined conditional probability paths:
\begin{equation}
    \mathcal{L}(\theta) = \mathbb{E}_{t \sim \mathcal{U}[0,1]} \mathbb{E}_{x \sim p_t} \left[ \| u_\theta(t, x) - u(t, x) \|^2 \right].
\end{equation}

\textbf{Improving flow matching for biomolecular structure prediction}: We introduced several key enhancements to make flow matching more suitable for probabilistic structure prediction: 

\begin{itemize}
    \item Informative priors: Flow matching offers greater flexibility in choosing the prior distribution functional form, and we reason that more appropriate prior distributions will better capture the underlying data structure. In NP3, we introduce a physics-motivated, globular polymer prior that preserves the distance structure among linked atoms and residues belonging to the same chain~(\cref{fig:np3_overview}A). This informative prior is efficiently implemented by relaxing random atom configurations with a limited number of Langevin dynamics iterations using an energy model with harmonic connectivity terms (\nameref{supp:si}, \cref{algo:a3:Globular}). 
    \item  Optimal transport structure permutation: By incorporating optimal transport principles, flow matching can create simpler flows that are more stable to train and lead to faster inference~\citep{tong_improving_2023,kornilov_optimal_2024,liu_flow_2022}.  We introduce a simulation-free symmetry correction module to straighten the conditional flow trajectories that connect the prior samples and ground truth structures. The symmetry correction module permutes equivalent entities, then permutes local atom indices while preserving the underlying chemical graph structures. 

    \item Vector field reparameterization: Instead of directly fitting the vector field $u$, we predict the denoised coordinates and estimate the vector field following an optimal rigid structure alignment. Following our past work~\citep{qiao2024state}, we also apply a rigid alignment against the previous-step structure estimate to improve trajectory continuity (\nameref{supp:si}, \cref{algo:a2:training}). 
\end{itemize}

More details on model training and inference can be found in \cref{algo:a1:sampling,algo:a2:training}. Altogether, these contributions led to two main advantages: (1) a dramatic reduction in the number of integrator steps needed to sample from the model, leading to improved inference efficiency; and (2) alleviating the need for expensive diffusion rollouts~\citep{abramson_accurate_2024} before each optimizer step, which both substantially speeds up and simplifies the procedure for training the main model and confidence modules. 

We have implemented several architectural enhancements to support these contributions.

\textbf{Model architecture}: Leveraging components from both its predecessors and recent structure prediction methods~\citep{abramson_accurate_2024,bryant_structure_2024}, we adopt an encoder-decoder architecture that separates the tasks of anchor-level conditioning and atom-level structure generation (\cref{fig:supply_arch}). Anchors are selected with a prespecified budget from all atoms, with priority given to residue backbones and ligand atoms. Paired multiple sequence alignment (MSA)~\cite{bacon_multiple_1986,bryant_improved_2022} features obtained via an improved pairing algorithm are assigned to these anchor atoms. The encoder comprises embedding layers to project all atom, bond and stereochemistry features; a hierarchical MSA Module; and AF3-style PairFormer blocks. The decoder consists of flow blocks using a diffusion transformer~\citep{peebles_scalable_2023} (DiT) architecture with additional geometric bias and modern normalization layers and layer initialization techniques. Each flow block operates first on anchor atoms using dense attention and then on all heavy atoms using a linear-scaling sliding window attention~\citep{beltagy_longformer_2020}, both using bias terms from input topology and encoder representations. Refer to the Supplementary Information sections \ref{supp:s1:data} and \ref{supp:s2:arch} for further details.

\begin{figure}
    \centering
    \includegraphics[width=0.9\linewidth]{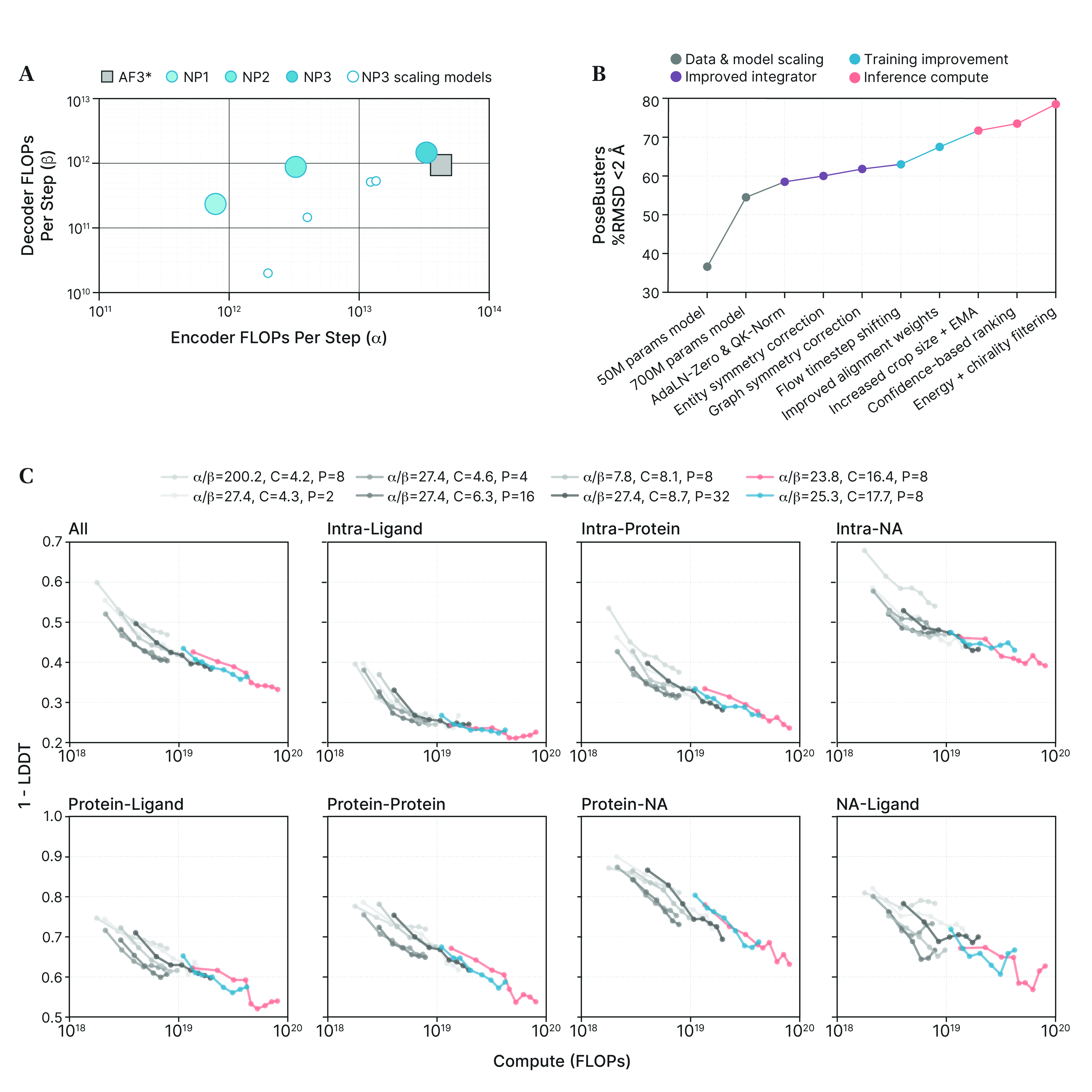}
    \caption{\textbf{NP3 pushes the frontier of structure prediction accuracy and training efficiency.} (\textbf{A}) Comparison of relative encoder/decoder capacity in terms of floating-point operations per second (FLOPs) among different methods. The asterisk indicates an estimate based on our reproduction of AF3. (\textbf{B}) Model scaling behavior across molecule and interface types as reported by the relation between the average validation set local distance difference test (LDDT)~\cite{mariani2013lddt} scores and the total training floating-point operation count (FLOPs). While intra-molecular prediction accuracy tends to saturate upon reaching a critical model size, inter-molecular interaction prediction benefits from jointly scaling the model and data up to the production model size. (\textbf{C}) Key model training and inference improvements included during the full course of model development and their impact on PoseBusters accuracy. $C$: Compute in GFLOPs; $P$: number of decoder replicas.}
    \label{fig:np3_performance}
\end{figure}

\begin{figure}
    \centering
    \includegraphics[width=0.6\linewidth]{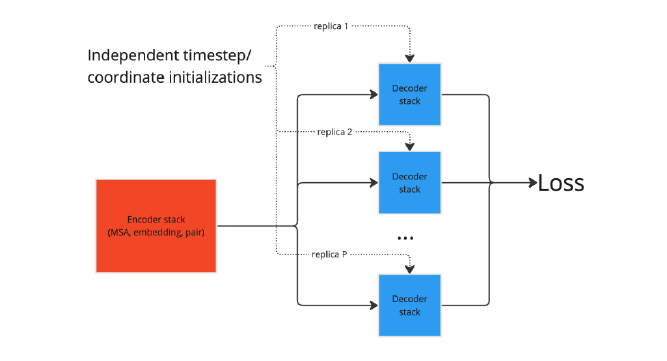}
    \caption{Schematics of decoder parallelism for training encoder-decoder structure prediction models.}
    \label{fig:supply_decoder}
\end{figure}

\textbf{Compute-optimal scaling}: The Chinchilla project~\citep{hoffmann_training_2022} introduced compute-optimal scaling laws, demonstrating that an optimal balance between model size and training data can be identified by extrapolating the Pareto frontier. However, applying these frameworks to structure prediction models presents unique challenges due to their (1) heterogeneous architectures; (2) the simultaneous training of conditioning and structure generation modules (\cref{fig:supply_arch}); (3) the use of decoder parallelism, where multiple independent coordinate initializations are passed to the decoder network within each training iterations to improve efficiency (\cref{fig:supply_decoder}); and (4) the vast number of modalities corresponding to each category of biomolecular interaction. To resolve these challenges, we reason that the total computational cost, measured in floating-point operations (FLOPs), serves as a more accurate indicator of model capacity than parameter count: 
\begin{equation}
    \text{Compute (FLOPs)} = (\alpha \cdot \beta \cdot P) \cdot D
\end{equation}

where encoder and decoder FLOPs are denoted as $\alpha$ and $\beta$, the number of training samples is denoted as $D$, and the number of decoder replicas per training iteration is denoted as $P$. To optimize the computational efficiency of our encoder-decoder architecture, we conducted a series of experiments to determine the ideal balance (\cref{fig:np3_performance}B) between the encoder-decoder capacity ratio $\alpha$ / $\beta$ and $P$. Our findings indicate that the compute-optimal frontier for NP3 is approximately ($\alpha$ / $\beta$=10, $P$=20). While intra-molecular prediction accuracy tends to saturate upon reaching a critical decoder size, inter-molecular interaction prediction continues to benefit from jointly scaling the model and data up to the production model size. We then marked the relative encoder/decoder capacity ($\alpha$ / $\beta$) among different methods, including our previous work, NP3 scaling models and the production 700M-parameter model, and an estimated result for AF3 based on our reimplementation (\cref{fig:np3_performance}A). 

\textbf{Model ablations}: We illustrate the impact of each modification included during the full course of model development on PoseBusters accuracy, categorized into data engineering, training loss improvements, improvements to the flow sampler, and better sample ranking protocols (\cref{fig:np3_performance}C). Key components that are found impactful include weighted interface-based training data sampling and cropping, AdaLN-Zero~\citep{peebles_scalable_2023}, QK-Norm~\citep{henry_query-key_2020}, and improved weighting scheme in rigid structure alignment that improves training (see \cref{algo:a2:training}), and the incorporation of clash and chirality penalties in conformer ranking.

\subsection*{Protein Structure and Interaction Accuracy }

\begin{figure}[h]
    \centering
    \includegraphics[width=0.6\linewidth]{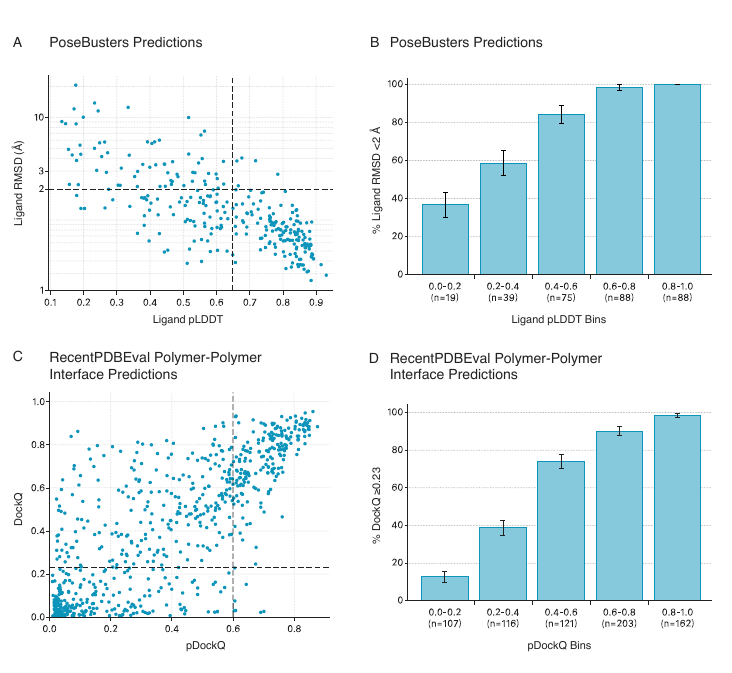}
    \caption{\textbf{Model confidence estimation.} (\textbf{A}) Scatter plot of ligand RMSD against ligand pLDDT and on PoseBusters. (\textbf{B}) PoseBusters ligand RMSD success rate statistics grouped into NP3 confidence percentiles. Prediction success rates are consistently higher for higher confidence prediction bins. (\textbf{C}) Scatter plot of DockQ score against the pDockQ score on NPBench protein-protein and protein-nucleic acid interfaces. (\textbf{D}) PPI DockQ success rate statistics grouped into NP3 confidence percentiles. Vertical lines on the scatter plot indicate the median prediction confidence. }
    \label{fig:np3_confidence}
\end{figure}

On the PoseBusters benchmark~\citep{buttenschoen_posebusters_2024}, NP3 achieves state-of-the-art accuracy in predicting protein-compound complexes (\cref{fig:np3_overview}B, \cref{tab:tb1_performance}, and \cref{fig:np3_confidence}). This benchmark evaluates two critical aspects that determine whether a model truly enables structure-based drug discovery: whether the predicted structures match experimental data to within the necessary accuracy (<2 \AA \space RMSD) and whether their molecular geometries are physically valid (PB-valid).

\begin{table}[h!]
\centering
\caption{\textbf{Quantitative model performance across biomolecule and interaction types}. Protein-ligand interaction prediction accuracy is evaluated on the PoseBusters-V2 dataset~\citep{buttenschoen_posebusters_2024}. Nucleic acids, covalent ligands, and protein prediction accuracies are evaluated using 1,143 chains or interfaces from low-homology, high-resolution, and deduplicated PDB structures released after 2023. }\label{tab:tb1_performance}\vspace{-27pt}
\includegraphics[width=0.999\linewidth]{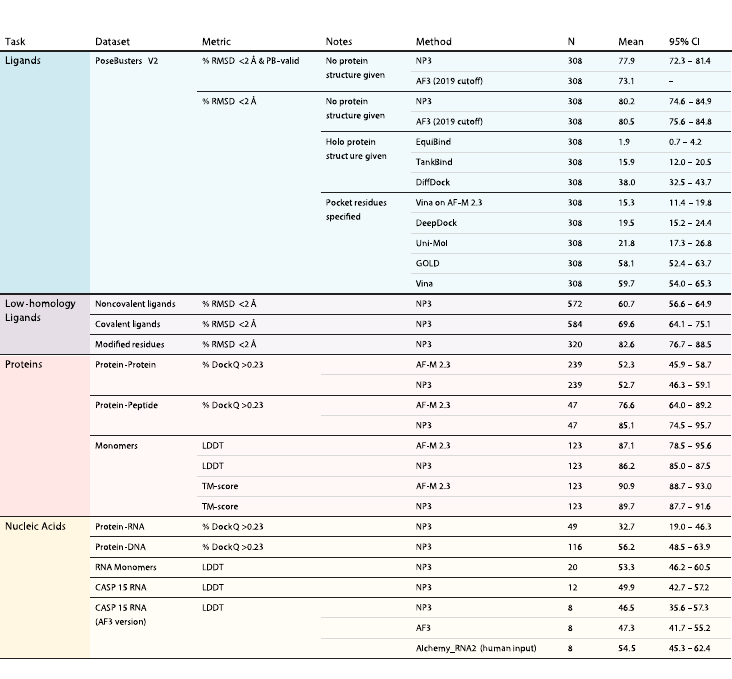}
\end{table}

Given only a protein sequence and a ligand's chemical structure, without prior knowledge of the protein's structure, binding site location, or ligand conformation, NP3 achieves a 78.4\% combined success rate across both metrics, outperforming AF3's 73.1\% (\cref{fig:np3_overview}B). On the coordinate error (\% RMSD < 2 \AA), NP3 achieves similar performance to AF3 (80.2\% versus 80.4\%, shown in \cref{tab:tb1_performance}).  NP3 similarly demonstrates substantial performance advantages over traditional docking methods that rely on experimental holo structures and explicitly defined pocket residues, including Vina (59.7\%), GOLD (58.1\%), and Uni-Mol (21.8\%)~\citep{eberhardt_autodock_2021,verdonk_improved_2003,zhou_uni-mol_2022}. Newer AI-based approaches, such as EquiBind, TankBind, and DiffDock~\citep{stark_equibind_2022,corso_diffdock_2022,tankbind}, which eliminate the need for explicitly defined pockets but still depend on reference protein structures, achieved lower success rates of 1.9\%, 15.9\%, and 38\%, respectively.  

Importantly, NP3 also achieves 98.8\% accuracy in predicting ligand stereochemistry (\cref{fig:np3_overview}B). This is critical for downstream drug design applications, as accurate stereochemical predictions are essential for optimizing ligand binding affinity and pharmacological activity, ensuring the development of effective and selective therapeutic compounds. 

To further evaluate NP3’s performance on nucleic acids, covalent ligands, and protein interaction accuracies, we introduced a new benchmarking suite, NPBench. NPBench addresses several limitations in existing benchmarks for structure prediction. Standard molecular docking benchmarks are typically limited to binary interactions (single target, single ligand) and therefore fail to capture the diversity of biomolecular interactions and stoichiometry inherent to molecular biology. Moreover, many of these benchmarks exhibit significant overlap with training structures, which can lead to an overestimation of model performance on novel targets. 

While AF3 provided a RecentPDBEval evaluation protocol that covers diverse molecular interactions, the associated code is not publicly available, limiting its accessibility for broader benchmarking efforts~\citep{abramson_accurate_2024}. In parallel, PLINDER introduced a comprehensive collection of protein-ligand interactions with stratified splits; however, it does not currently support nucleic acids, generalized stoichiometry, or holdout splits aligned with the time-based evaluation splits commonly used in modern structure prediction models~\citep{durairaj_plinder_2024}.

To address these gaps, NPBench comprises 1,143 chains or interfaces derived from low-homology, high-resolution, and deduplicated PDB structures released after 2023. Further details of this benchmark are described in the Methods section, and the code will be made publicly available to support reproducibility and future advancements in the field. 

On this benchmark, NP3 exhibited an overall strong performance across diverse biomolecular targets and interactions: 

\begin{itemize}
    \item \textbf{Protein monomers and protein-protein interactions (PPIs)}: NP3 demonstrated comparable performance in predicting protein-protein interactions, achieving a success rate of 52.7\% on low-homology interfaces with DockQ~\citep{mirabello_dockq_2024} scores greater than 0.23, closely matching AlphaFold2-Multimer v2.3 (AF2-M 2.3)~\citep{evans_protein_2022}, which achieved 52.3\%. For monomers, NP3 and AF2-M 2.3 also showed similar results, with success rates of 87.1\% and 86.2\%, respectively, as measured by LDDT. 
    \item \textbf{Protein-peptide interfaces}: On protein-peptide interfaces—where peptides are defined as standard or modified polypeptide chains with fewer than 20 amino acids—NP3 largely outperformed AF2-M, achieving an 85.1\% success rate compared to AF2-M’s 76.6\%. 
    \item \textbf{Noncovalent ligands}: NP3 achieved a success rate of 60.7\% (ligand RMSD <2 \AA) for noncovalent ligands in the evaluation set, compared to 80.2\% on the PoseBusters benchmark. Notably, the evaluation set consists of ligand-target interfaces where either the binding pocket or the ligand exhibits significant structural dissimilarity from the training data. These results demonstrate NP3's ability to generalize effectively to unseen targets and novel molecules, maintaining satisfactory performance under more stringent evaluation criteria. 
    \item \textbf{Covalent ligands and PTMs}: For covalent ligands, NP3 achieved a high success rate of 69.6\%, while for modified residues it reached 82.6\%. These results underscore NP3's ability to generalize to proteins with chemical modifications and covalent chemistry, making it particularly relevant for drug discovery applications targeting chemically modified proteins. 
    \item \textbf{Nucleic acids}: On CASP15 RNA targets~\citep{kryshtafovych_critical_2023,das_assessment_2023}, NP3 demonstrated comparable performance to AF3 (46.5\% versus 47.3\%, respectively) and slightly below that of Alchemy\_RNA2~\citep{chen_rna_2023} (54.8\%), which relies on explicitly curated human inputs. While AF3 conditions its predictions on RNA MSAs, NP3 uses RNA language models (LMs) and achieves similar performance—a noteworthy result, as MSAs are typically considered more effective than LMs for conditioning structure predictions. This outcome also highlights the potential of utilizing LM-based approaches in scenarios where RNA MSAs are unavailable or impractical to generate. For protein-DNA interactions, NP3 achieved accuracy comparable to protein-protein interactions (56.2\% versus 52.7\%, respectively) and lower, though reasonable, accuracy for protein-RNA interactions (32.7\%). 
\end{itemize}

A strong correlation was also observed between NP3’s model-estimated confidence and true prediction accuracy, highlighting its ability to effectively gauge the reliability of its outputs (\cref{fig:np3_confidence}). Confidence for ligand predictions was quantified using the pLDDT score, while polymer-polymer interface predictions were assessed using the newly introduced pDockQ score (see SI, \cref{supp:s5:sample}). To further analyze the relationship between confidence and accuracy under a cutoff that defines prediction successful-ness, prediction success rates were aggregated into percentiles based on model-estimated confidence. Results showed that higher confidence predictions achieved substantially better outcomes. Specifically, for predictions in the top 50\% confidence subset, NP3 achieved a 96.7\% RMSD success rate on the PoseBusters benchmark and a 93.1\% DockQ success rate on the recent Protein Data Bank (PDB) evaluation dataset.

\subsection*{Computational Efficiency and Prediction Speed}

\begin{figure}[t]
    \centering
    \includegraphics[width=0.95\linewidth]{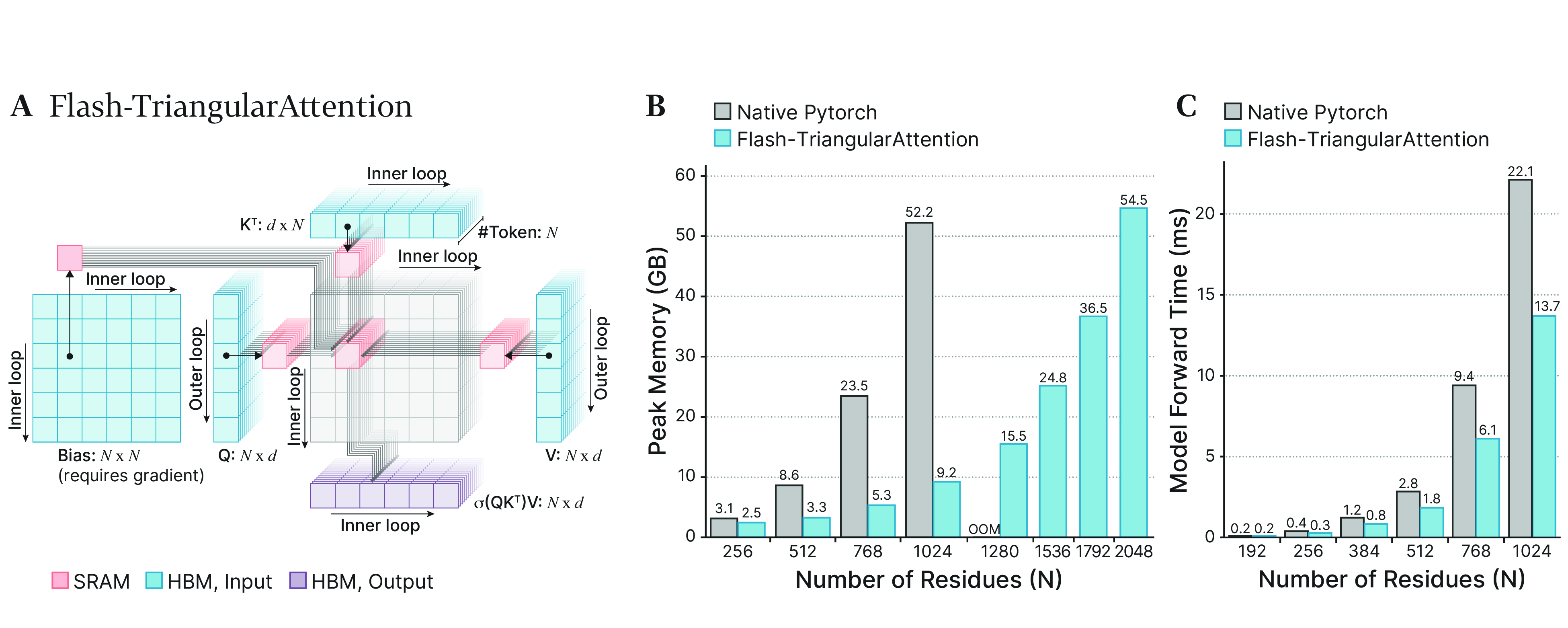}\vspace{-10pt}
    \caption{\textbf{Illustration of Flash-TriangularAttention kernel and experimental comparison on the peak memory usage and the inference time.} (\textbf{A}) The workflow for Flash-TriangularAttention. The main goal is to reduce peak memory usage, which is the bottleneck of triangular attention. 
    Our implementation avoids this explicit broadcast, which is required while using other memory-efficient implementations, thus enabling longer crop size training.  (\textbf{B}) Peak memory usage comparison between Flash-TriangularAttention and PyTorch built-in SDPA. Our implementation significantly reduces peak memory usage for longer residues, enabling samples with residues exceeding $1024$ to run on one H100 GPU (80GB). (\textbf{C}) the inference time of the Flash-TriangularAttention and Pytorch built-in SDPA. The figure shows that our implementation significantly reduces inference runtime—for example, by $38\%$ at a residue of $1024$. }
    \label{fig:np3_flashtriattn}
\end{figure}

To address the computational challenges of molecular structure prediction, we introduced Flash-TriangularAttention, a novel attention mechanism optimized for efficient training and inference (\cref{fig:np3_overview}A and \cref{fig:np3_flashtriattn}). The method leverages implicit bias broadcasting within the kernel during the $\mathbf{QK}^T + \mathbf{Bias}$ operation, which eliminates unnecessary bias replication, significantly reducing memory overhead and enabling scalability for large structures.   

In more detail, the triangular attention mechanism requires a gradient-tracked $\mathbf{Bias}$ term. The query ($\mathbf{Q}$), key ($\mathbf{K}$), and value ($\mathbf{V}$) tensors each have the shape $(B * H, N, N, d)$, and the $\mathbf{Bias}$ tensor has the shape $(B * H, N, N)$, where $B$ is the batch size, $H$ is the number of attention heads, $N$ is the crop size, and $d$ is the head dimension. In a naive approach, during the $\mathbf{QK}^T + \mathbf{Bias}$ operation, $\mathbf{Bias}$ is broadcasted from $(B * H, N, N)$ to $(B * H, N, N, N)$. Current optimized solutions, such as FlashAttention~\citep{dao_flashattention_2022} and PyTorch’s scaled dot-product attention (SDPA) and FlexAttention~\citep{li_flexattention_2025}, either do not support gradient-tracked $\mathbf{Bias}$ or require explicit bias broadcasting which dramatically increases peak memory usage. To train models with larger structural crop sizes, we seek to avoid this broadcasting step.  

Flash-TriangularAttention leverages implicit bias broadcasting within the kernel during the $\mathbf{QK}^T + \mathbf{Bias}$ operation. For each channel, the bias is loaded once from the bias matrix during the forward pass, while the backward pass accumulates gradients into a shared bias matrix. This eliminates unnecessary bias replication, significantly reducing memory overhead and enabling scalability for large values. Our Triton kernel implementation, a language and compiler for writing highly efficient custom Deep-Learning primitives, enhances efficiency, allowing for training on larger sequences and crop sizes while staying within GPU memory constraints. Experimental results underscore the effectiveness of Flash-TriangularAttention (\cref{fig:np3_flashtriattn}). Peak memory usage decreases by $5\times$ compared to a naive implementation as crop size increases, enabling the model to scale effectively, while the forward pass inference time is improved by $50\%$ on average. These improvements make NP3 capable of processing large, complex structures without compromising computational efficiency.

Building on these advancements, NP3 delivers predictions $15\times$ faster than inference timing statistics reported by AF3 (\cref{fig:np3_overview}B). Unlike AF3~\citep{abramson_accurate_2024} which requires approximately six A100-minutes per prediction, NP3 generates results in approximately $30$ seconds using a single L40S GPU. This efficiency is also attributed to removing the expensive trunk recycling operations and cutting the number of sampling steps to $40$ by leveraging optimal-transport flow samplers (\cref{fig:supply_arch}).

\subsection*{Ligand-induced Conformational Change Prediction}

While NP3’s advancements in prediction accuracy and speed represent significant progress, standardized benchmarks remain critical for evaluating its utility in drug discovery applications. One such benchmark is the assessment of accuracy in predicting correct apo (non-ligand-bound) and holo (ligand-bound) structures. Accurate conformational predictions, both globally and locally around the binding site, are of acute importance in drug discovery by assisting in identifying allosteric binding sites and optimizing interactions for improved efficacy and selectivity.  

Several groups have discussed AlphaFold’s limitations in predicting complete conformational ensembles of proteins of interest~\citep{lazou_predicting_2024,sala_modeling_2023,saldano_impact_2022,meller_predicting_2023,riccabona2024assessing}. Even so, there is a lack of standard benchmarks to quantitatively measure the prediction accuracy of proteins with induced conformational changes.

\begin{figure}[tp]
    \centering
    \includegraphics[width=0.9\linewidth]{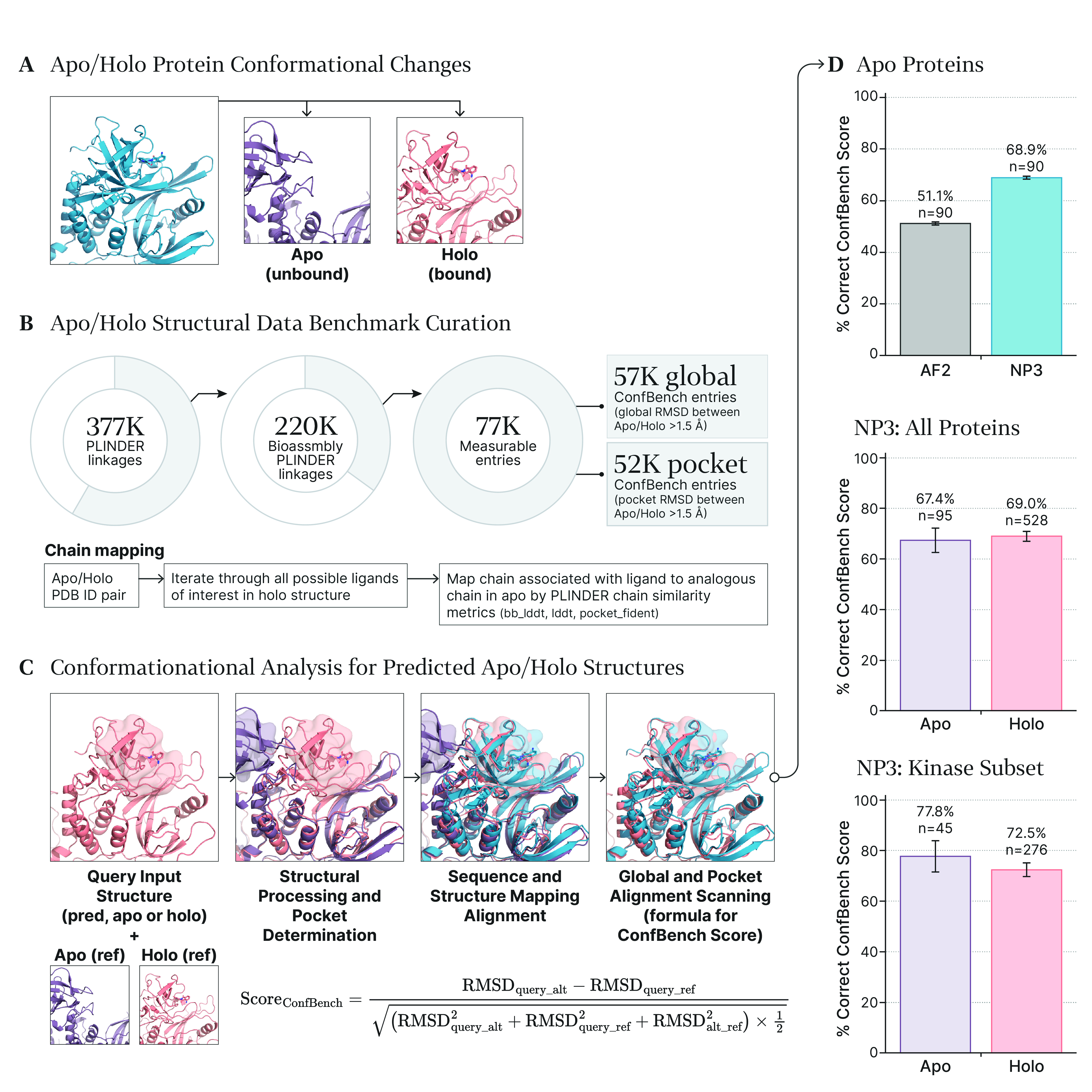}
    \caption{\textbf{ConfBench conformational prediction}. (\textbf{A}) Schematic illustration of types of conformational changes of interest. (\textbf{B}) Benchmark target statistics. 57k global and 52k pocket rearrangement entries with meaningful motions and biologically relevant ligands are identified. (\textbf{C}) The ConfBench scoring protocol. (\textbf{D}) Conformational prediction success rate statistics comparison. A correct conformational prediction is defined as a ConfBench score greater than 0. NP3 outperforms AF2-M 2.3  in predicting pocket conformations on apo targets, while maintaining similar success rates on apo targets and holo targets.}
    \label{fig:np3_confbench}
\end{figure}

To address this need we developed ConfBench, a conformational benchmark that systematically evaluates ligand-induced conformational changes across the proteome and assesses model conformational biases by comparing predictions against reference apo and holo states (\cref{fig:np3_confbench}). ConfBench consists of a rigorously curated selection of chain and ligand-specified apo/holo PDB pairs that exhibit measurable conformational differences in one or more of the following ways: global RMSD > 1.5\AA, pocket alpha carbon RMSD > 1.5\AA, or pocket alpha carbon plus sidechain heavy atom RMSD > 1.5\AA. Additionally, we propose a conformational scoring function that is a measure of conformational accuracy which is insensitive to absolute variance between reference structures and also agnostic of structural accuracy of conformationally irrelevant domains.  

To enable the PDB-wide measurement of conformational change down to an atomic level of accuracy, a stringent alignment protocol was required prior to RMSD. Starting from PLINDER~\citep{durairaj_plinder_2024}, we implemented an exhaustive, bidirectional search of all linkages, examining all possible chain combinations and distinct ligand binding sites in multi-liganded complexes. Optimal chain mapping was comprehensively validated with PLINDER chain similarity metrics, namely pocket sequence identity (pocket\_fident), chain-wise local distance difference test (LDDT), and chain-wise backbone local distance difference test (bbLDDT).  

We introduce a conformational pairing protocol for complexes involving the binding of multiple ligands or cofactors, adopting a definition such that the paired apo structure is always defined based on a ligand of interest selected from a holo structure regardless of alternative binders or other contextual information that is less relevant.  Ligand-of-interest-centric chain mapping enables conformational accuracy measurements of structures beyond monomers while still preserving their overall global structural context, allowing for the conformational scoring of both allosteric and orthosteric binders. 

For apo/holo pairs that have conformational changes greater than 1.5\AA, i.e. changes that are measurable and relevant, NP3 outperforms AF2-M in conformational prediction success rates. NP3 outperformed AF2-M for 54 unique linkages with a correctness rate of 51.9\% of predictions, compared to 29.6\% of AF2 predictions. This is true across the board, whether that be global conformational changes, pocket alpha carbon atom conformational changes, or pocket including sidechain conformational changes. Compared to context-unaware methods like AF2-M, this result demonstrates the potential of using all-atom structure prediction algorithms to capture conformational variations induced by ligand binding and change in physiological conditions. 

NP3 achieves consistent performance across apo and holo structures: Across all evaluated proteins, NP3 achieved correctness rates of 67.4\% for apo structures (N\_linkages = 95) and 69.9\% (N\_linkages = 5280) for holo structures. To further evaluate performance within a high-value therapeutic target class, the benchmark dataset was refined to focus on kinases. Within this subset, NP3 achieved correctness rates of 77.8\% for apo structures and 72.5\% for holo structures. 

There are some limitations: (1) imperfect ConfBench scores for NP predictions and (2) apo structures used for scoring do not always imply a fully ligand-free structure.  We recognize that ligand-free structures in the PDB are relatively limited~\citep{durairaj_plinder_2024,feidakis_ahoj-db_2024}; we expect more work to enable a conformational benchmark that is diverse, biologically relevant, and leakage free.

\section*{Discussion}

Accurately determining protein-ligand structures and their conformational distributions remains a significant challenge, especially at the fine scale required for drug discovery. Along these lines, one notable contribution of NP3 is its high accuracy on the PB-valid metrics that assess the prediction of physically plausible structures. This capability is essential for generating meaningful insights and avoiding unreliable results, as physically realistic predictions directly impact the translational success of computational models in drug discovery. NP3’s high accuracy in predicting PTMs further represents a step forward in modeling biological processes such as signaling, immune response, and protein folding~\citep{lane2023protein}. These targets are historically underexplored in structure-based drug discovery due to experimental challenges, where NP3 can open new opportunities for drug targeting. 

In modeling specific conformations, particularly apo and holo states, NP3 also demonstrates improved accuracy compared to AF-M 2.3. Notably, NP3 shows a similar fraction of correct conformational predictions between its predictions on apo targets and holo targets. These precise predictions, such as the successful modeling of kinase conformations, are critical for optimizing lead compounds and deepening the understanding of target protein behavior in structure-based drug design, and we anticipate the evaluation framework of this work to be a basis for fairly measuring and improving conformation modeling ability. 

Finally, NP3’s ability to deliver predictions in seconds sets a new benchmark in computational efficiency. This speed advantage enables researchers to virtually screen extensive chemical libraries and iterate on ligand designs with significantly greater scalability. Moreover, NP3 achieves this efficiency while maintaining compatibility with standard hardware, reducing computational resource requirements and making advanced structure prediction more accessible. 

By combining rapid, high-fidelity prediction with data-driven functional insights, NP3 establishes a robust framework for structural biology, drug discovery, and protein engineering. This advancement positions NP3 as a pivotal tool for the next generation of therapeutic discovery and molecular design.

\section*{Methods }
\textbf{Structure Prediction Benchmarks}: We evaluated NP3’s structural prediction performance on three benchmarks—the PoseBusters-V2 dataset~\citep{buttenschoen_posebusters_2024} for protein-ligand interaction prediction accuracy, NPBench for diverse interaction types and stoichiometry on low-homology structures, and CASP15 for RNAs~\citep{das_assessment_2023}.  

\textbf{Baselines}: We conducted comparisons across multiple benchmark datasets to evaluate performance. For the PoseBusters ligand dataset~\citep{buttenschoen_posebusters_2024}, we compared NP3 to traditional docking methods that require holo receptor structures, including Vina~\citep{eberhardt_autodock_2021} and GOLD~\citep{verdonk_improved_2003}, as well as machine learning-based methods with similar holo structure requirements, such as EquiBind~\citep{stark_equibind_2022}, TankBind~\citep{lu_tankbind_2022}, and DiffDock~\citep{corso_diffdock_2022}. Additionally, we benchmarked against structure prediction methods that use only sequence and graph information, including AF3. Baseline numbers for these methods were derived from Abramsom et al (Extended Data Table 1)~\citep{abramson_accurate_2024}. 

For protein targets from the dataset of recent PDB structures (NPBench), NP3’s performance was compared to AlphaFold-Multimer 2.3 using ColabFold~\citep{mirdita_colabfold_2022}. Five seeds were generated per target, and the top-ranked conformers were scored using both NP3 metrics and the "multimer" metric provided by AlphaFold-Multimer~\citep{evans_protein_2022}. 

For CASP15 RNA targets, NP3 was benchmarked against AF3 and Alchemy\_RNA2~\citep{chen_rna_2023}. Baseline metrics for these methods were sourced from the AF3 publication, which originally evaluated eight CASP15 RNA targets~\citep{abramson_accurate_2024}. With all CASP15 RNA targets now publicly available, NP3’s average RNA LDDT scores were reported across both the original subset and the complete dataset, with no significant differences in performance between subsets. 

\textbf{Evaluation Criteria}: To evaluate model performance across this diverse set of biomolecules and structures, we utilized several well-established metrics: 

\begin{itemize}
    \item RMSD <2 \AA \space success rate: The fraction of predictions with pocket-aligned ligand RMSD below 2 Angstrom, a widely used metric for assessing structural prediction accuracy. 
    \item PB-valid success rate: The proportion of PoseBusters predictions that pass physical validity checks implemented in PoseBusters (PB).
    \item Local distance difference test (LDDT): An alignment-free metric for assessing the accuracy of local structures across cutoff all biomolecules. A 25 \AA \space inclusion cutoff is used for DNAs and RNAs, and a 15 \AA \space is used for proteins. 
    \item DockQ >0.23 success rate: The fraction of DockQ score greater than 0.23 across all interfaces for evaluation, a metric for evaluating the accuracy of polymer-polymer interaction predictions using acceptable thresholds from CAPRI~\citep{janin2003capri}. 
\end{itemize}

\textbf{Conformational Benchmarks}: To develop the ConfBench benchmark, which assesses conformational accuracy, all PLINDER~\citep{durairaj_plinder_2024} linkages were filtered for entries with only experimental structures from the PDB. These linkages were further distilled to curate a dataset of protein-ligand systems exhibiting measurable conformational changes between apo and holo states (\cref{fig:np3_confbench}A and B). Systems with ``lock-and-key" binding, where the protein structure remains largely unchanged upon ligand binding, or those with minimal allosteric shifts, were excluded. This step ensures the benchmark focuses on clear, quantifiable ligand-induced structural transitions that are significant enough to measure and therefore evaluate prediction accuracy and enable effective model performance assessment. 

We next employed a systematic approach to identify precise chain-level correspondences between apo and holo protein structures—a critical metric for reliably evaluating ligand-induced conformational shifts in multichain, multi-liganded oligomeric systems:

\begin{enumerate}
    \item Extract chain information from PLINDER’s structured identifiers; 
    \item For each protein chain with an associated ligand chain, identify all possible mappings between holo protein chain and linked apo protein chains;
    \item Rank these mappings based on quantitative structural metrics: 
        \begin{itemize}
            \item Pocket\_fident: Sequence identity within the binding pocket 
            \item LDDT: Local distance difference test 
            \item bb\_LDDT: Backbone local distance difference test 
        \end{itemize}
\end{enumerate}

To evaluate structural predictions, we employed a scoring formula designed to quantify the similarity between predicted (query) and ground truth (reference) structures. The score is calculated as: 
 
\begin{equation}
    \text{score} = \frac{\text{RMSD}_{\text{query\_alt}} - \text{RMSD}_{\text{query\_ref}}}{\sqrt{\left(\text{RMSD}^2_{\text{query\_alt}} + \text{RMSD}^2_{\text{query\_ref}} + \text{RMSD}^2_{\text{alt\_ref}}\right)} \times \frac{1}{2}}
\end{equation}

where: 

\begin{itemize}
    \item score = 1 when the query RMSD is 0, indicating the prediction is identical to the reference state
    \item score = -1 indicates the prediction is identical to the linked alternative state
    \item score = 0 indicates the prediction is equally distant from the reference as the apo/holo states.
\end{itemize}

Between 0 and 1 the score scales smoothly, reflecting the degree of similarity to the reference state. 

\textbf{Code Availability}: NPBench is made openly accessible and permissively licensed under BSD-3c to enable benchmarking arbitrary structure prediction methods on the recent PDB bioassemblies dataset, Posebusters, and new dataset extensions. The NPBench repository is accessible at \url{https://github.com/iambic-therapeutics/np-bench}.

\section*{Acknowledgments}
This work used computational resources provided by the National Energy Research Scientific Computing Center (NERSC), under Contract No. ERCAP0029432.

\printbibliography

\begin{refsection}

\clearpage
\beginsupplement
\section[SI]{Supplementary Information}
\label{supp:si}

\subsection{Data and Input Pipelines}\label{supp:s1:data}

\textbf{Molecular Topology Featurization}: The initial structural features passed to the NP3 inference pipeline are a complete representation of the molecular topology. These input features include: 

\begin{itemize}
    \item Atom-wise features: atom types and residue types. Residue types include a one-hot encoding of the 1-letter residue name for atoms from standard amino acids and nucleotides, and a one-hot encoding for any other residues.
    \item Atom-pair features: bond orders, and bond orientations:
    \begin{itemize}
        \item For a bonded atom pair $(i, j)$, we evaluate the bond orientations by first constructing a local frame based on two nearest bonded atoms $(ik, il)$ using the generalized Gram-Schmidt procedure described in \citep{qiao2024state}, and assign the bond orientation feature as the direction of the displacement vector between atom $i$ and atom $j$ expressed in the assigned local frame.
    
        \item We assign the bond order to 0 and bond orientation features to (0, 0, 0) for any non-bonded atom pairs. 
        
        \item Disulfides, ionic bonds, hydrogen bonds, and halogen bonds are excluded from featurization and solely inferred by the model. 
    \end{itemize}
\end{itemize}

Heavy atoms of a bioassembly are split into two levels of treatment: atoms and anchors. Only the anchors and anchor pairs among them are being processed by the encoder block to produce conditionings, while all atoms are passed through the flow blocks to produce denoised coordinates of the entire assembly. Given a fixed context size budget, anchors are selected as follows: 

\begin{enumerate}
    \item Biopolymer backbones are prioritized. We label all C$\alpha$ atoms of amino acid residues and C1' atoms for nucleic acids as anchors, unless the budget is filled in which case we perform a sampling without replacement;
    \item Then we perform additional sampling without replacement from all atoms from ligands, non-standard residues, and PTMs;
    \item If any budget is left, we perform atom sampling from the remaining standard protein, DNA, and RNA residues until the budget limit is reached.
\end{enumerate}

\textbf{Evolutionary sequence features}: NP3 additionally leverages conditioning from MSAs, protein language models, and nucleic acid language models. For training, we follow the MSA generation protocol from AF3, combining jackhmmer MSAs~\citep{eddy_accelerated_2011} on uniref30~\citep{suzek_uniref_2015}, reduced BFD, and uniref100; and hhblits MSAs~\citep{steinegger_hh-suite3_2019} for BFD~\citep{steinegger_protein-level_2019}. We do not include RNA MSAs. We use a 3B-parameter language model for proteins (ESM-2)~\citep{lin_evolutionary-scale_2023} and a 650M-parameter language model for RNAs (RiNalMo)~\citep{penic_rinalmo_2024}.

We introduce a compactified MSA pairing algorithm to capture cross-chain coevolutionary genetics while efficiently utilizing a fixed token context window of at most 16,384 sequences. The method proceeds by first sorting sequences from all databases with decreasing sequence similarity w.r.t the query sequence, then pairing sequences by taxonomical ID~\citep{bryant_improved_2022}, and finally backfilling the blank spaces with unpaired monomer sequences.  

\textbf{Data filtering and preprocessing pipeline}: For structures obtained from the PDB, we generally follow the mmcif-level filtering protocol that has been described for AlphaFold 3~\citep{abramson_accurate_2024}, with the following minor differences: 
\begin{itemize}
    \item We skip structures where either the asymmetric unit cif file size or the bioassembly cif file size exceeds 20 MB;
    \item We do not need to remove leaving group atoms based on information about the residue in the CCD;
    \item We do not limit the structures passed to the training pipeline to 20 chains.
\end{itemize}

To obtain correct connectivity and bond types, we make use of the information provided in the CCD for non-standard residue names via ParmED’s `all\_residue\_template\_match'~\citep{shirts_lessons_2017}. Furthermore, we read the bonds present in the `struct\_conn' section in the cif file for the asymmetric unit, which is usually missing in cif files for bioassemblies. We then identify matching atom pairs in the bioassemblies and add the corresponding bonds.

\begin{table}[h]
\centering
\caption{Comparison of data processing pipeline details.}
\label{tab:data_processing_comparison}
\begin{tabular}{p{7cm}cc}
\toprule
\textbf{Training Data Processing Step} & \textbf{AF3} & \textbf{NP3 (Ours)} \\ 
\midrule
Keep structures of reported resolution of 9 \AA\ or less & \checkmark & \checkmark \\ \midrule
Remove structures with 300 or more polymer chains & \checkmark & \checkmark \\ \midrule
Remove polymer chains containing fewer than 4 resolved residues & \checkmark & \checkmark \\ \midrule
Remove hydrogen atoms & \checkmark & \checkmark \\ \midrule
Remove clashing chains (those with $>$30\% of atoms within 1.7 \AA\ of an atom in another chain) & \checkmark & \checkmark \\ \midrule
Remove atoms outside of the CCD code's defined set of atom names & \checkmark & \checkmark \\ \midrule
Remove leaving atoms for covalent ligands & \checkmark & Not needed \\ \midrule
Filter out protein chains with consecutive C$\alpha$ atoms greater than 10 \AA\ distance & \checkmark & \checkmark \\ \midrule
Truncate closest 20 chains relative to random token for large bioassemblies & \checkmark & Keep all chains \\ \midrule
For structures from crystallography, remove crystallization aid molecules & \checkmark & \checkmark \\ 
\bottomrule
\end{tabular}
\end{table}

\subsection{Model architecture and inference}\label{supp:s2:arch}
In \cref{algo:a1:sampling}, we outline NP3’s structure sampling procedure. CNFs integrate a learned velocity field to transform a prior distribution and sample new data. Our conformer sampling protocol iteratively refines structures through time-conditioned decoding, rigid-body structure between integrator steps alignment to reduce discretization errors, and graph–preserving atom permutation corrections. In practice, we also apply a timestep shifting trick using a polynomial transform $t^* = t^{1.15}$, which is found beneficial to reduce exposure bias~\citep{esser_scaling_2024}.

\begin{algorithm}[!tp]
\caption{Sampling from NPv3 with symmetry-corrected flow. CFM: Conditional flow matching.}\label{algo:a1:sampling}
\begin{algorithmic}[1]
\Function{NPv3SampleStructure}{NPv3Model, $\mathbf{c}$, $T=100$} \hfill {\color{blue}$\triangleright$ T: Number of integrator steps}
    \State $\mathbf{x}_0 \gets \mathrm{MultichainPolymerPriorSample}(\mathbf{c})$ \hfill {\color{blue}$\triangleright$ Sample initializations from the physics-inspired prior}
    \State $\mathbf{x}_1^{\text{last}}$ $\gets \mathbf{x}_0$, $\Delta t \gets 1/T$, $t \gets 0$
    \State $\mathbf{z}_\mathrm{atom}, \mathbf{z}_\mathrm{anchor}, \mathbf{z}_\mathrm{pair-anchor}, \mathbf{z}_\mathrm{pair-local} = \mathrm{NPv3Encoder(\mathbf{c})}$
    \For{$\mathrm{timestep}$ \textbf{in} $0$ \textbf{to} $T-1$}
        \State $t_\mathrm{next} \gets t + \Delta t$
        \State $\mathbf{x}_1^{\text{pred}} \gets \text{NPv3Decoder}(\mathbf{x}_t, t | \mathbf{z}_\mathrm{atom}, \mathbf{z}_\mathrm{anchor}, \mathbf{z}_\mathrm{pair-anchor}, \mathbf{z}_\mathrm{pair-local})$
        \State $\mathbf{r}_{\text{opt}}, \mathbf{t}_{\text{opt}} \gets \text{GetKabschTransform}(\mathbf{x}_1^{\text{pred}}, \mathbf{x}_1^{\text{last}}, \mathbf{w}_\mathrm{align})$ \hfill {\color{blue}$\triangleright$ Global SE(3)-superposition to improve continuity}
        \State $\mathbf{x}_1^{\text{pred}} \gets \mathbf{x}_1^{\text{pred}} \cdot \mathbf{r}_{\text{opt}} + \mathbf{t}_{\text{opt}}$
        \State {$\mathbf{x}_1^{\text{pred}} \gets \text{OptimalGraphPermutation}(\mathbf{x}_1^{\text{pred}}, \mathbf{x}_1^{\text{last}})$} \hfill {\color{blue}$\triangleright$ Atom permutation correction using graph isomorphism}
        \State $\mathbf{x}_1^{\text{last}}$ $\gets \mathbf{x}_1^{\text{pred}}$ 
        \State $\mathbf{x}_{t_\mathrm{next}} \gets \text{CFMSampleStep}(\mathbf{x}_1^{\text{pred}}, \mathbf{x}_t, t, t_\mathrm{next})$ \hfill $\mathrm{CFMSampleStep}(x_1, x_t, t, t_\mathrm{next}) = t_\mathrm{next} \cdot x_1 + (1 - t_\mathrm{next}) \cdot \frac{x_t - t \cdot x_1}{1 - t}$
        \State $t \gets t_\mathrm{next}$
    \EndFor
    
    \State \Return $x_1$
    
\EndFunction
\end{algorithmic}
\end{algorithm}

When sampling initial coordinates from the prior, we incorporate constraints at multiple scales. Bonded connectivity ensures that linked atoms form coherent polymer-like chains; Entity-level and residue-level scale constraints guide atomic positions to cluster in a manner consistent with coarse-grained structural motifs; and a global spherical confinement maintains a compact, globular arrangement.

\begin{algorithm}
\caption{NPv3 main model training iteration.}\label{algo:a2:training}
\begin{algorithmic}[1]
\Function{NPv3TrainingIter}{NPv3Model, $\mathbf{x}_1^\mathrm{ref}$, $t$, $\mathbf{c}$, $\sigma_\mathrm{data} = 16.0Å$, $\epsilon = 0.01$}  \hfill {\color{blue}$\triangleright$ $\mathbf{x}_1^\mathrm{ref}$: Ground truth structure}
    \State $\mathbf{x}_0 \gets \mathrm{MultichainPolymerPriorSample}(\mathbf{c})$ \hfill {\color{blue}$\triangleright$ Sample initializations from the physics-inspired prior}
    \State $\mathbf{x}_0 \gets \text{OptimalEntityPermutation}(\mathbf{x}_0, \mathbf{x}_1^{\text{ref}})$ \hfill {\color{blue}$\triangleright$ Apply entity permutation on prior sample}
    \State $\mathbf{x}_t \gets (1-t) \cdot \mathbf{x}_0 + t \cdot \mathbf{x}_1^\mathrm{ref}$
    \State $\mathbf{x}_1^{\text{pred}} \gets \text{NPv3Model}(\mathbf{x}_t, t^* | \mathbf{c})$, \quad $t^* = t^{1.15}$ \hfill {\color{blue}$\triangleright$ Encoder + decoder forward pass}
    \State $\mathbf{r}_{\text{opt}}, \mathbf{t}_{\text{opt}} \gets \text{GetKabschTransform}(\mathbf{x}_1^{\text{pred}}, \mathbf{x}_1^{\text{ref}}, \mathbf{w}_\mathrm{align})$ \hfill {\color{blue}$\triangleright$ Global SE(3)-superposition using backbone trace}
    \State $\mathbf{x}_1^{\text{ref}} \gets \mathbf{x}_1^{\text{ref}} \cdot \mathbf{r}_{\text{opt}} + \mathbf{t}_{\text{opt}}$
    \State {$\mathbf{x}_1^{\text{pred}} \gets \text{OptimalGraphPermutation}(\mathbf{x}_1^{\text{pred}}, \mathbf{x}_1^{\text{ref}})$}  \hfill {\color{blue}$\triangleright$ Atom permutation correction using graph isomorphism}
    \State $\mathbf{x}_1^{\text{ref}}$ $\gets \text{STOP\_GRAD}(\mathbf{x}_1^{\text{ref}})$
    \State $w(t) \gets 1 / {(\epsilon + (1-t) \sigma_\mathrm{data} )}$ \hfill {\color{blue}$\triangleright$ Loss weighting following \citep{lee_improving_2024}}
    \State $\mathcal{L} = w(t) \cdot \mathrm{PseudoHuberLoss(\mathbf{x}_1^{\text{pred}}, \mathbf{x}_1^{\text{ref}}) + w_2 \cdot \mathrm{SmoothLDDT(\mathbf{x}_1^{\text{pred}}, \mathbf{x}_1^{\text{ref}})}} + w_3 \cdot \mathrm{FAPE}(\mathbf{x}_1^{\text{pred}}, \mathbf{x}_1^{\text{ref}})$
    \State Backpropagate on $\mathcal{L}$
\EndFunction
\end{algorithmic}
\end{algorithm}

\begin{algorithm}[H]
\label{alg:prior}
\caption{Sampling from a Globular Polymer Prior via Short Langevin Dynamics}\label{algo:a3:Globular}
\begin{algorithmic}[1]
\Require $X_0 \in \mathbb{R}^{N_{\text{atoms}} \times 3}, dt=0.25, \text{sphere\_r}, \text{res\_r}=4.0, \text{ent\_r}=10.0$
\Require Matrices $S_{\text{bond}}, S_{\text{entity}}, S_{\text{residue}}$ define neighbor and group structure
\For{$i=1\to 64$}
   \State $d_{\text{bond}} \gets (S_{\text{bond}}X_0) - X_0$
   \State $d_{\text{entity}} \gets (S_{\text{entity}}X_0) - X_0$
   \State $d_{\text{res}} \gets (S_{\text{residue}}X_0) - X_0$
   \State $\text{drift} \gets 2\,d_{\text{bond}} + \frac{d_{\text{entity}}}{\text{ent\_r}^2} + \frac{d_{\text{res}}}{\text{res\_r}^2} - \frac{X_0}{\text{sphere\_r}^2}$
   \State $X_0 \gets X_0 + dt \cdot \text{drift} + 2\sqrt{dt}\,\epsilon \quad\text{with}\;\epsilon \sim \mathcal{N}(0,I)$
\EndFor
\State \textbf{return} $X_0$
\end{algorithmic}
\end{algorithm}

Starting from random atomic positions, the algorithm iteratively applies drift updates derived from these constraints and adds small thermal noise. After a brief relaxation period, the resulting coordinates serve as a well-structured initialization that respects chemical connectivity, coarse geometric organization, and global compactness. This initialization naturally breaks inter-chain symmetry for copies of the same biopolymer. During training, at the end of the initialization we apply a greedy permutation of the prior entities such that the nearest neighbor entity index of each entity maximally agrees with the nearest neighbors derived from the ground truth.

\begin{figure}[!tp]
    \centering
    \includegraphics[width=0.8\linewidth]{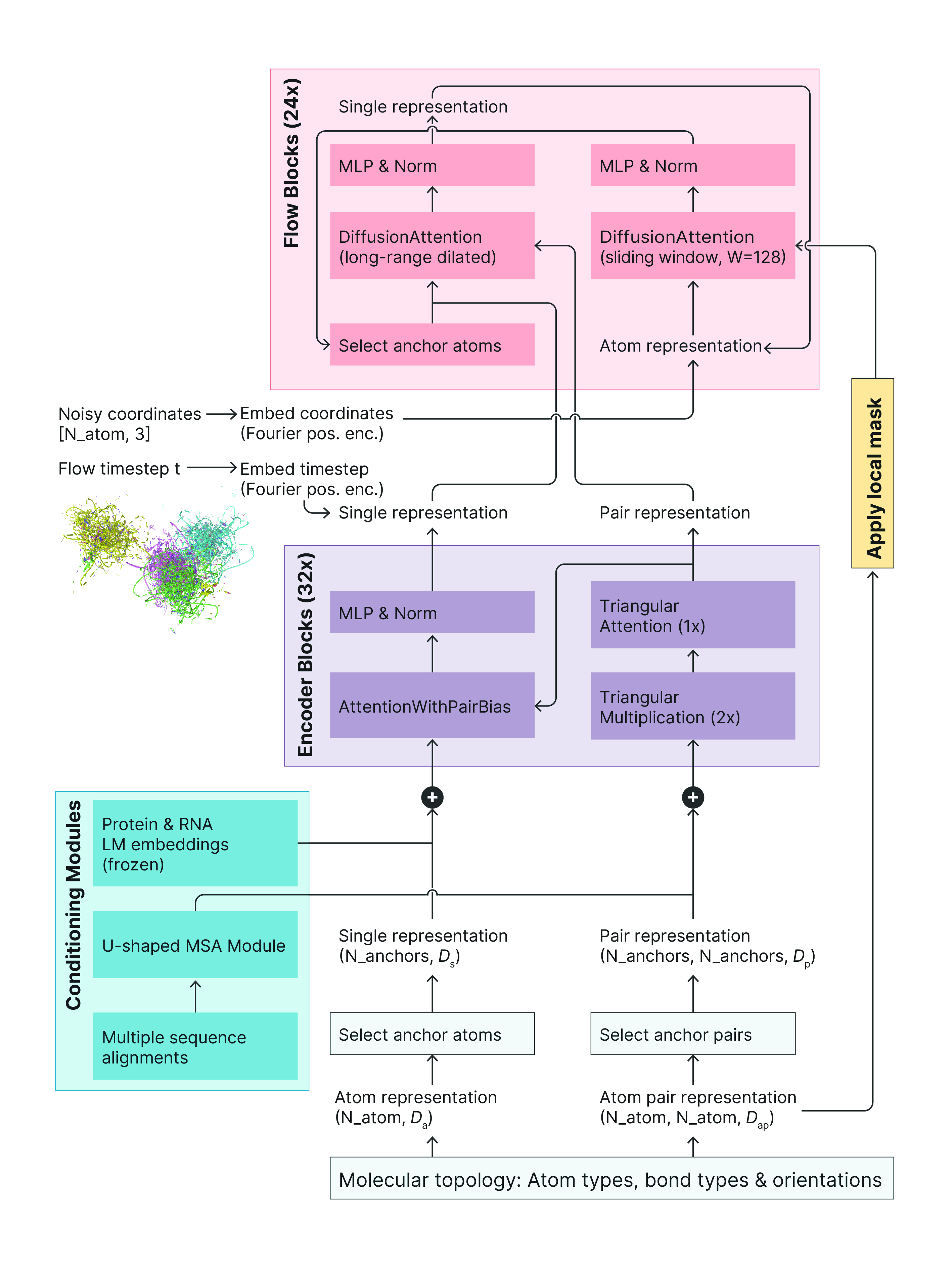}
    \caption{Architecture details. $D_\mathrm{a}$: atom representative latent dimension; $D_\mathrm{s}$: single representation latent dimension; $D_\mathrm{p}$: pair representation latent dimension; $D_\mathrm{ap}$: sliding window atom pair representation latent dimension.}
    \label{fig:supply_arch}
\end{figure}

In \cref{fig:supply_arch}, we provide a detailed view of the NP3 model architecture. The production model uses the following configuration: 350M encoder parameters, 350M decoder parameters, 50M confidence module parameters. We highlight the following \textbf{architecture details:} 

\begin{itemize}
    \item \textbf{Standard block details}: We use a single representation embedding size of 384, pair representation embedding size of 128, atom representation embedding size of 256, and atom pair representation size of 128. For MLP blocks, we use a hidden size that is 4 times the embedding size similar to the design choices made in Llama-2~\citep{touvron_llama_2023}. RMSNorm~\citep{zhang_root_2019} is used for all MLP blocks; LayerNorm~\citep{ba_layer_2016} without learnable bias is used for all MSA module blocks, pair bias attention, triangular attention, and triangular multiplication blocks.
    \item \textbf{U-Shaped MSA Module}: NP3 uses an MSA Module consisting of 6 blocks, each with the architecture described in \citep{abramson_accurate_2024}. The MSA latent dimension progressively scales from 32 to 512 with a multiplier of 2 after each block, optimizing memory usage with minimal representation capacity tradeoff.
    \item \textbf{Sliding window attention with sparse pair bias}: We employ an efficient implementation of sliding window attention compatible with pairwise attention bias, maintaining linear asymptotic inference time and memory scaling with respect to the number of heavy atoms. The atom pair features (bonds and orientations) are projected by a linear layer and collated to the sliding window mask. The biased sliding window masks are shared across all decoder replicas when using decoder parallelism during training.
\end{itemize}

\subsection{Model training}\label{supp:s3:training}

\textbf{Dataset Splitting and sampling: }
\begin{itemize}
    \item NP3 is trained on all PDB structures deposited \textbf{before September 1, 2020}, along with additional synthetic datasets. The cutoff date is chosen to ensure no overlap with evaluation set structures. We process training samples based on contiguous and spatial cropping centered around each to ensure maximal training data mixture diversity. In our implementation, we always sort the chain numbers based on the minimum distance to the chain or interface of interest before applying contiguous cropping.
    \item \textbf{Weighted sampling} of structure crops is employed to ensure inclusion of underrepresented molecule classes (e.g., nucleic acids, ligands). We mainly follow AF3 for clustering all chains and interfaces including a 40\% maximum similarity threshold for sequence clustering, with the notable difference of using 60\% Tanimoto similarity cutoff for ligand clustering rather than the 100\% CCD identity cutoff employed in AF3.
    \item \textbf{Data augmentation} strategies include spatial translation, rotation, MSA row-wise subsampling, and 15\% residue masking rate for language model embeddings, and 15\% masking rate for atom type, bond order, and bond orientation features.
\end{itemize}

Several synthetic datasets are used to augment model training and improve generalization, with examples drawn from computational predictions: 

\begin{itemize}
    \item Protein monomer distillation: With 30\% probability, we sample a structure from the OpenProteinSet dataset of OpenFold~\citep{ahdritz_openfold_2024} predictions (N=270,000).
    \item Disordered distillation set for protein multimers. With 1\% probability, we sample a structure from NP2 predictions of bioassemblies containing more than one polymer chain, with the motivation to reduce hallucination for intrinsically disordered regions. These samples were generated by NP2 with full-length sequence inputs and template coordinates guidance, for PDB structures with 50 or more contiguous missing residues (N=2,801).
    \item OrbNet conformers: With 2\% probability, we sample a single molecular conformer. These conformers are obtained from a comprehensive dataset of small molecule conformations refined at the $\omega$B97X-D3/Def2-TZVP level of DFT theory, expanded from OrbNet Denali training data (N=12,706,496).~\citep{christensen_orbnet_2021}
\end{itemize}

\textbf{Model weight initialization}: Within all attention layers, we adopt QK-normalization~\citep{henry_query-key_2020} to stabilize model training. For all PairFormer blocks, we employ Zero-residual initialization motivated by recent works~\citep{jumper_highly_2021,zhao2021zero}. For Flow blocks, we employ AdaLN-Zero initialization~\citep{peebles_scalable_2023} to both the attention and MLP layers, with scale and shift parameters dependent on timestep embedding. 

\textbf{Optimization Details}: In \cref{algo:a2:training}, we outline the procedure of a single optimization step for training NP3. Following Lee et al.~\citep{lee_improving_2024}, we found that pseudo-Huber loss combined with SmoothLDDT introduced in AlphaFold 3 yields the most stable training. Additional loss terms include a distance-normalized Frame-Aligned Point Error (FAPE) and a distance-geometry loss to ensure accurate local and global structural alignments~\citep{qiao2024state}.

We modified the alignment weights $\mathbf{w}_\mathrm{align}$ used in the weighted Kabsch-Umeyama algorithm~\cite{kabsch_solution_1976,umeyama_least-squares_1991}, which we found important to reduce numerical variations in late-stage training: 

\begin{itemize}
    \item Atom weight = 10.0 for the ligand of interest, when the training structure is generated based on a contiguous or spatial crop around a ligand-polymer interface;
    \item Atom weight = 1.0 for C$\alpha$ and C1” atoms, and all rest ligand atoms;
    \item Atom weight = 0.0 for all side-chain atoms.
\end{itemize}

\textbf{Optimization algorithm and training stages}: The NP3 production model was trained on 64 H100 GPUs for 24 days.  Training is carried out using PyTorch FSDP~\citep{ansel_pytorch_2024} under BF16 automatic mixed precision. Over the full course of model training, we progressively scale the structure crop size to gradually capture longer contexts. The number of decoder replicas P is tuned for each stage: 

\begin{itemize}
    \item Maximum cropping size for first 80\% training iterations: 384 anchors or 3072 atoms, P=32;
    \item For 10\% of training iterations: 512 anchors or 4096 atoms, P=32
    \item For 6\% of training iterations: 768 anchors or 8192 atoms, P=24;
    \item For the last 4\% of training iterations: 1024 anchors or 12800 atoms, P=12.
\end{itemize}

Activation checkpointing is employed for all training stages with anchor crop size larger than 384. Note that training on structure crops of 1024 anchors or greater sizes was only possible when using the Flash-TriangularAttention kernel, otherwise we observe GPU OOM.

\subsection{Training Strategy for Confidence Module}\label{supp:s4:confidence}

\begin{algorithm}[H]
\caption{Schematics of NP3 Confidence Module Training}
\label{alg:confidence_training}
\begin{algorithmic}[1]
\Require $\text{NPv3Model}$, ground truth $\mathbf{x}_\text{ref}$, conditions $\mathbf{c}$
\For{each iteration $n$}
    \State Sample $t \sim \text{Uniform}[0, 1)$ \hfill {\color{blue}$\triangleright$Sample timestep from noise schedule}
    \State Sample noised structure $\mathbf{x}_t \leftarrow \text{SampleFromPrior}(t | \mathbf{c})$
    \State Compute denoised structure $\mathbf{x}_\text{denoised} \leftarrow \text{NPv3Model}(\mathbf{x}_t)$
    \State Compute flow matching loss: $\mathcal{L}_\text{flow}(\mathbf{x}_\text{ref}, \mathbf{x}_\text{denoised})$
    \State $\text{LDDT} \leftarrow \text{LDDT}(\mathbf{x}_\text{denoised}, \mathbf{x}_\text{ref})$  \hfill   {\color{blue}$\triangleright$Compute per-atom LDDT score}
    \State $\text{DistanceError} \leftarrow |\mathrm{Dmap}(\mathbf{x}_\text{denoised}) - \mathrm{Dmap}(\mathbf{x}_\text{ref})|$    \hfill   {\color{blue}$\triangleright$Compute anchor distance map errors}
    \State Pass $\mathbf{x}_\text{denoised}$ through confidence module and extract \text{confidence\_logits}
    \State $\mathcal{L}_\text{confidence} \leftarrow \text{CrossEntropy}(\text{confidence\_logits}, \{\text{LDDT}, \text{DistanceError}\})$  \hfill {\color{blue}$\triangleright$Compute confidence loss}
    \State $\mathcal{L}_\text{total} \leftarrow \mathcal{L}_\text{flow} + \mathcal{L}_\text{confidence}$  \hfill  {\color{blue}$\triangleright$ Update model with total loss}
    \If{$n \mid 10$}
        \State $\mathbf{x}_\mathrm{sampled} \leftarrow \text{NPv3SampleStructure}(\text{NPv3Model}, \mathbf{c}, T=10)$  \hfill  {\color{blue}$\triangleright$Apply rollout with limited steps}
        \State Pass $\mathbf{x}_\text{sampled}$ through confidence module and extract \text{confidence\_logits}
        \State $\mathcal{L}_\text{total} \leftarrow \mathcal{L}_\text{total} + \mathcal{L}_\text{confidence}(\mathbf{x}_\mathrm{sampled}, \mathbf{x}_\text{ref}, \text{confidence\_logits})$ 
    \EndIf
\EndFor
\end{algorithmic}
\end{algorithm}

NP3 provides two types of confidence scores:  

\begin{itemize}
    \item Predicted local distance difference test (pLDDT) for each heavy atom;
    \item Predicted distance error (pDE) for each pair among anchors.
\end{itemize}

In NP v1-2 and AF3, confidence module training is performed by sampling bioassembly structures via full diffusion Stochastic Differential Equations (SDE) at predefined intervals N, followed by optimization steps to minimize deviations between model-estimated errors and true distance errors. We note that AF3 uses N=1 (perform diffusion rollout and optimize at every iteration), while NP1, NP2, and NP3 use N=10. 

Analysis of preliminary models revealed limited sensitivity of confidence scores to sampled conformations, despite good correlation between accuracy and confidence scores over diverse targets. We reasoned that the insensitivity of confidence scores stems from under-training and insufficient paired conformational data for the same molecular topology. 

The improved training algorithm integrates additional optimization steps into the standard flow-matching procedure. This approach involves sampling noisy conformations, denoising them through the model, and optimizing confidence scores based on cross-entropy loss with calculated structural errors. Noise levels are varied, producing multiple hypotheses for the same molecular topology, enabling robust confidence estimation. The algorithm thus ensures efficiency while minimizing the need for costly resampling or simulation-based inference. In the fine-tuning stages, we also incorporate a InfoNCE~\citep{infonce} loss commonly adopted in contrastive learning to encourage the model to discern minor prediction quality differences across multiple conformations.

\subsection{Aggregated Confidence Metrics and Sample Ranking}\label{supp:s5:sample}

Aggregated confidence metrics are provided for entities and interfaces: 

\begin{itemize}
    \item pLDDT for atomic-level predictions, averaged against atoms from the entity of interest.
    \item pDockQ for self- or inter-molecular interactions between a chain pair. 
\end{itemize}
    
pDockQ is calculated based on the following formula:
\begin{equation*}
    \mathrm{pDockQ} = (\mathrm{iRMS\_scaled\_lr} + \mathrm{iRMS\_scaled\_sr} + \mathrm{Predicted\_Fnat}) / 3
\end{equation*}

where:
        \begin{itemize} 
        \item iRMS\_scaled\_lr = 1 / (1 + (mean(pDE$^2$)) / 8.5$^2$)
        \item iRMS\_scaled\_sr = 1 / (1 + (mean(pDE$^2$)) / 1.5$^2$)
        \item Predicted\_Fnat = mean(pDE  < 5.0)
        \end{itemize}

\textbf{Ranking Methodology}: Sample ranking leverages adjusted pLDDT and pDockQ values. Scores are penalized for steric clashes, ensuring physically plausible configurations. 

\begin{itemize}
    \item For a ligand of interest, LG: Score = pLDDT(LG) – 1000 * (is\_clash + is\_chirality\_violation)
    \item For chain A of interest: Score = pDockQ(A, A)
    \item For chain pair (A, B) of interest: Score = pDockQ(A, B).
\end{itemize}

\subsection{Evaluation datasets and benchmarking (NPBench)}\label{supp:s6:NPBench}
\textbf{Curation of the recent PDB dataset} is loosely based on the strategy introduced by AF3 Supplemental Information Section 6.1-6.2~\citep{abramson_accurate_2024} The dataset construction started by taking all 10,192 PDB entries deposited between 2022-05-01 and 2023-01-12. We filter to non-NMR entries with resolution better than 4.5A. To choose the most representative sample from these candidates with strong protection against overlap from training set, we perform a PDB clustering on chain/ligand/interface index for all structures released before 2023-01-12, with a maximum similarity cutoff of 40\% sequence identity for polymers and 0.6 Tanimoto similarity for ligands and PTMs. Then, for the recent structures, we:

\begin{itemize}
    \item Include monomers from novel clusters (not covered by PDB clusters used during training);
    \item Include polymer-polymer interfaces if at least one of the polymers is in novel clusters;
    \item Include polymer-peptide interfaces if the receptor chain is in novel clusters;
    \item Include all polymer-ligand, ion, and metal interfaces with additional labels: \newline is\_ligand\_unseen\_CCD, is\_plsite\_unseen\_cluster, is\_covalent, is\_modified\_residue. 
    \item Exclude all exotic cases not covered by criteria above. In particular, we exclude all branch entities such as multi-residue glycosylations in this benchmark release as there is no clear consensus on their representations in the context of structure prediction model inputs.
\end{itemize}

We then created the benchmark input mmCIF files by removing solvents and crystallography artifacts from the original samples. Finally, we deduplicate the dataset by only keeping the first occurrence of chain or interface from each cluster and removing all samples on which ColabFold inference did not complete within 1 GPU hour, yielding 1,143 evaluation targets in total.

\textbf{Benchmarking metrics }
The following metrics were used to assess the accuracy of the predictions on the benchmarking datasets: 
\begin{itemize}
    \item Pocket-aligned ligand RMSD: We used a rigorous alignment protocol to compute the pocket-aligned ligand RMSD which is described as follows:
    \begin{itemize}
        \item We define the reference pocket as all C$\alpha$ atoms within 10  of the ligand atoms in the ground-truth structure. If the ligand-polymer interface is not specified (as in Posebusters benchmarking), the C$\alpha$ atoms that belong to the chain with the most residues in the reference pocket are selected for alignment.
        \item The same set of C$\alpha$ atoms in the predicted structure are selected based on the residue indices and the optimal chain mapping between the predicted and the ground-truth structures obtained using DockQ v2. If the optimal chain mapping is not provided (as in Posebusters benchmarking), we will align each chain in the predicted pocket, defined in the same way as the reference pocket, and find the minimal RMSD.
        \item We use PyMOL~\citep{delano_pymol_2002} to align the selected C$\alpha$ atoms and all ligand atoms in the predicted structure to those in the ground-truth structure with zero refinement cycles.
        \item The pocket-aligned ligand RMSD is computed using RDKit~\citep{Landrum2016RDKit2024_09_1} CalcRMS between the aligned ligand and the reference ligand structures.
    \end{itemize}
    
    \item DockQ v2: The DockQ scores are computed for all polymer-polymer interfaces defined in the Recent PDB Evaluation Set using DockQ v2~\citep{mirabello_dockq_2024} package.
    \item Generalized RMSD: This is computed for each ligand-polymer interface defined in the Recent PDB Evaluation Set as follows:
    \begin{itemize}
        \item We first find the optimal chain mapping between the predicted and the ground-truth structures using DockQ v2.
        \item We use the same algorithm as described above to compute the pocket-aligned ligand RMSD for a given ligand-polymer interface.
        \item For structures with multiple identical ligands, we perform an exhaustive search over all pairwise permutations between the predicted and reference ligands (up to 10 ligands) and find the minimal pocket-aligned ligand RMSD. This leads to $N^2$ pocket-aligned RMSD calculations where $N$ is the number of ligands corresponding to the same CCD code in a ligand-polymer interface.
    \end{itemize}
    \item TM-score: This is computed using the USalign package. The TM-scores for both entire protein and each chain are calculated using the optimal chain mapping obtained from DockQ.
    \item LDDT (local distance difference test): We compute the all-atom and backbone-only LDDT scores for both entire structure and each chain in the predictions with a custom implementation.
\end{itemize}

\subsection{Conformational change benchmarking (ConfBench)}\label{supp:s7:ConfBench}

\subsubsection{ConfBench Dataset Curation}
The development of ConfBench required establishing a systematic protocol for identifying and validating apo-holo protein structure pairs that exhibit measurable conformational changes. Here we detail the technical implementation of our structure pair identification pipeline and the specific criteria used for quality control.

\textbf{Structure Pair Database Integration}:
The curation pipeline was built on PLINDER's structured database schema, leveraging its system\_id indexing that encodes both PDB identifiers and chain information in a standardized format (PDB\_\_model\_\_protein.chain\_\_ligand.chain). We developed a two-pass search algorithm that first identifies potential apo-holo relationships through system\_id queries, then validates these relationships through detailed structural analysis.

\textbf{Chain Mapping Protocol}: 
For structures with multiple chains, our mapping algorithm evaluates all possible chain combinations using a hierarchical sorting approach. The best mapping is selected by first sorting by $\texttt{pocket\_fident}$ (binding site sequence conservation), followed by $\texttt{lddt}$ (overall structural similarity) and $\texttt{bb\_lddt}$ (backbone conformation) as successive tiebreakers. This prioritizes binding site similarity while still considering global structural features when discriminating between similar candidates.

\textbf{Conformational Change Validation}:
Structure pairs were required to exhibit at least one of the following:
\begin{itemize}
   \item Global RMSD > 1.5\AA
   \item Pocket C$\alpha$ RMSD > 1.5\AA
   \item Pocket all-atom RMSD > 1.5\AA
\end{itemize}

\subsubsection{ConfBench Scoring Protocol}
The ConfBench scoring protocol introduces several key technical advances to address quantifying the quality of prediction of ligand-induced conformational changes:
\begin{itemize}
   \item A robust protocol for identifying and mapping binding site residues between structures that accounts for numbering discontinuities and chain breaks
   \item A multi-level structural comparison approach that evaluates both global and local conformational changes
   \item A novel scoring function that enables fair comparison across protein systems with varying magnitudes of conformational change
\end{itemize}

\textbf{Pocket Detection and Mapping}:
Pocket residues were identified using a distance-based approach in PyMOL, selecting protein residues within 10\AA \space of the ligand of interest. To ensure consistent pocket mapping between structures, we implemented a robust sequence alignment-based protocol that accounts for potential numbering discontinuities and chain breaks. The mapping algorithm uses local sequence alignment with the Bio.Align.PairwiseAligner module, prioritizing sequence identity while handling insertions and deletions. Pocket mappings were validated by requiring at least 50\% of the reference pocket residues to be successfully mapped to maintain structural context.

\textbf{RMSD Calculations and Structural Alignment}: Three levels of structural comparison were implemented:
\begin{itemize}
   \item Global alignment using all C$\alpha$ atoms
   \item Pocket-specific alignment using only pocket C$\alpha$ atoms
   \item Complete pocket alignment including both backbone and sidechain heavy atoms
\end{itemize}

Each alignment was performed using PyMOL's align command with cycles=0 to prevent local optimization bias. For multi-chain structures, chain mapping was determined through a hierarchical approach - first attempting sequence-based matching, then falling back to spatial proximity to the ligand if necessary.

\textbf{Conformational Score Calculation}: 
The conformational scoring function was designed to be:
\begin{itemize}
   \item Symmetric with respect to the reference structures
   \item Normalized to account for varying magnitudes of conformational change
   \item Invariant to global rigid body movements
\end{itemize}

For each alignment type i (global, pocket, pocket+sidechains), the score is calculated as:

\begin{equation}
    \mathrm{score\_apo\_i} = \frac{\mathrm{RMSD\_holo} - \mathrm{RMSD\_apo}}{\sqrt{\frac{1}{2} (\mathrm{RMSD\_holo}^2 + \mathrm{RMSD\_apo}^2 + \mathrm{RMSD\_ref}^2)}}
\end{equation}
\begin{equation}
    \mathrm{score\_holo\_i} = \frac{\mathrm{RMSD\_apo} - \mathrm{RMSD\_holo}}{\sqrt{\frac{1}{2} (\mathrm{RMSD\_apo}^2 + \mathrm{RMSD\_holo}^2 + \mathrm{RMSD\_ref}^2)}}
\end{equation}

where RMSD\_ref is the RMSD between reference apo and holo structures, and RMSD\_apo/RMSD\_holo are RMSDs between the query structure and respective references. This formulation ensures scores are bounded and comparable across different protein systems regardless of their absolute conformational differences.

\subsubsection{Additional NP3 and AF2-M results}

When limiting to "big" ConfBench wins (i.e. ConfBench scores >> 0), NP3 achieves 47.7\% of wins above ConfBench score of 0.5, whereas AF2-M achieves 32.22\%.

\begin{figure}[h]
    \centering
    \includegraphics[width=0.7\linewidth]{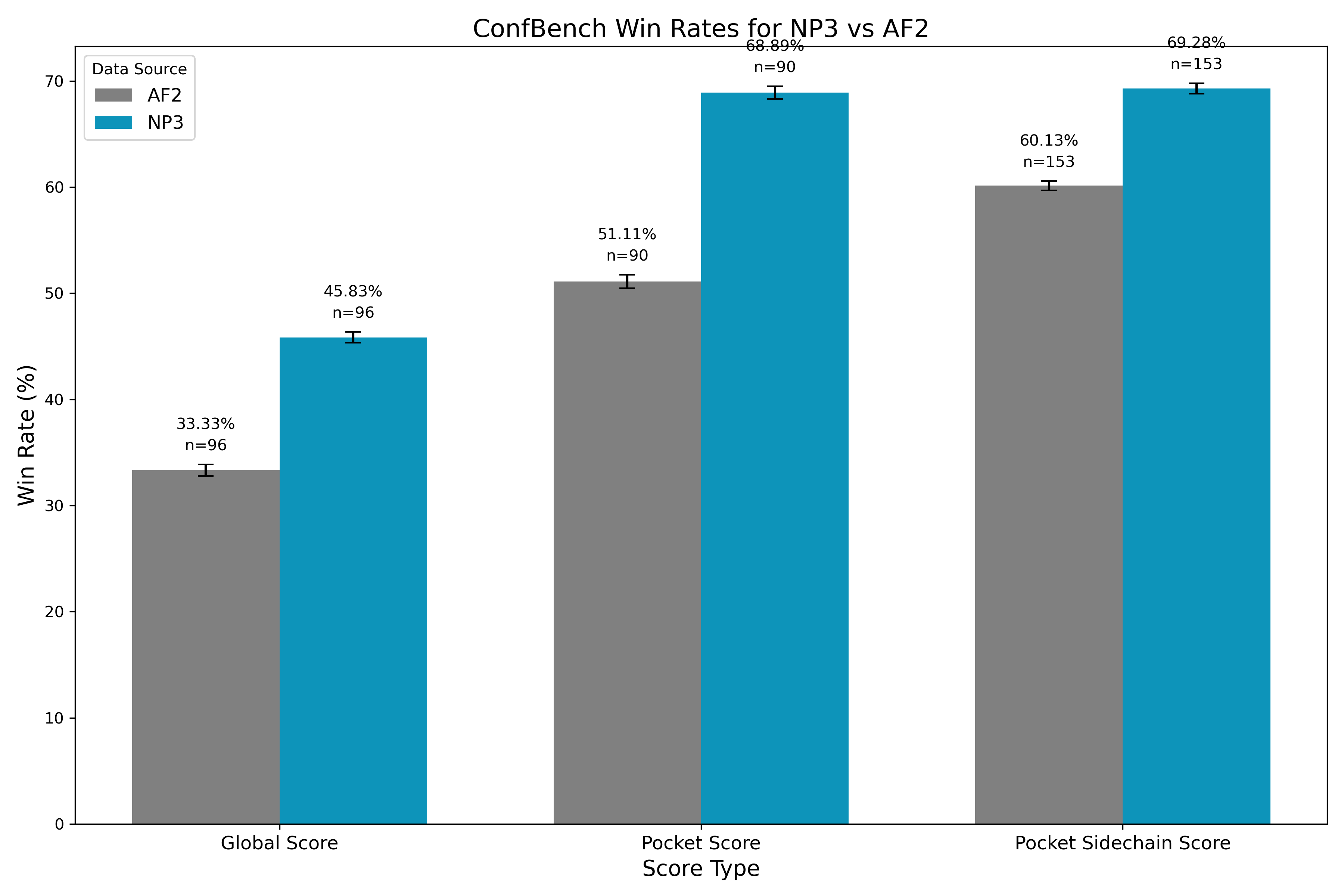}
    \caption{Win rates for all ConfBench score types on query conformation = apo linkages where reference RMSD > 1.5 Å. A ConfBench win is defined as query structure score > 0. }
    \label{fig:enter-label}
\end{figure}

Detailed distributions of prediction accuracy are presented through histograms of ConfBench scores and subsequent kernel density estimation plots comparing AlphaFold2-Multimer and NP3 performance, providing comprehensive visualization of the complete prediction landscape beyond aggregate statistics.

For additional clarity, histograms of NP3 raw ConfBench score distribution data are shared below, detailing the magnitude of win rates for the apo and holo dataset discussed in the main text.

\begin{figure}[h]
    \centering
    \includegraphics[width=0.99\linewidth]{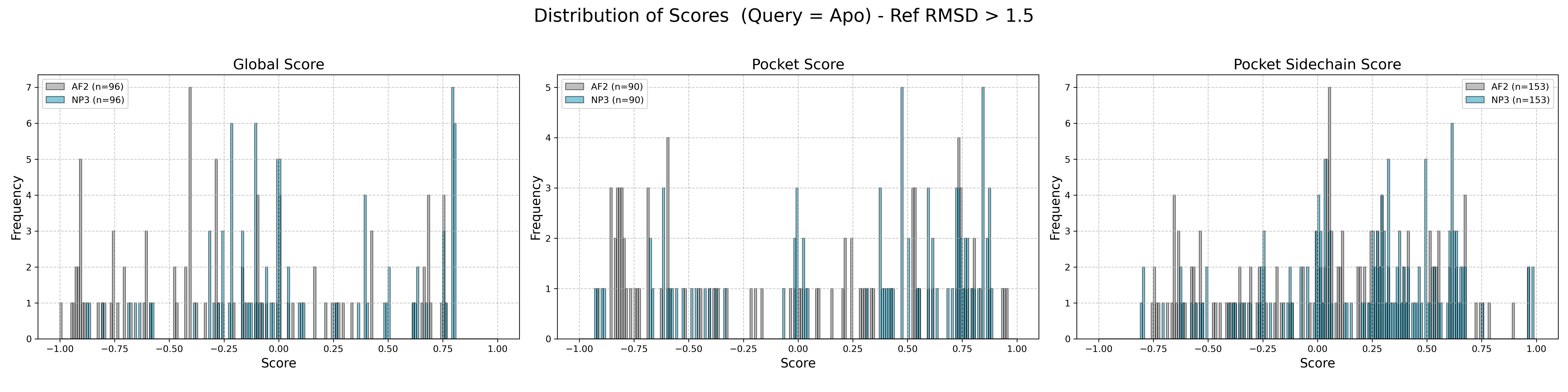}
    \caption{Distribution of ConfBench score values for NP3 and AF2 multimer models with query conformation = apo and a reference RMSD > 1.5 \AA. The histograms display the frequency of scores across all metrics (global, pocket, and pocket-sidechain) within the range of -1 to 1.}
    \label{fig:enter-label}
\end{figure}

\begin{figure}[h]
    \centering
    \includegraphics[width=0.99\linewidth]{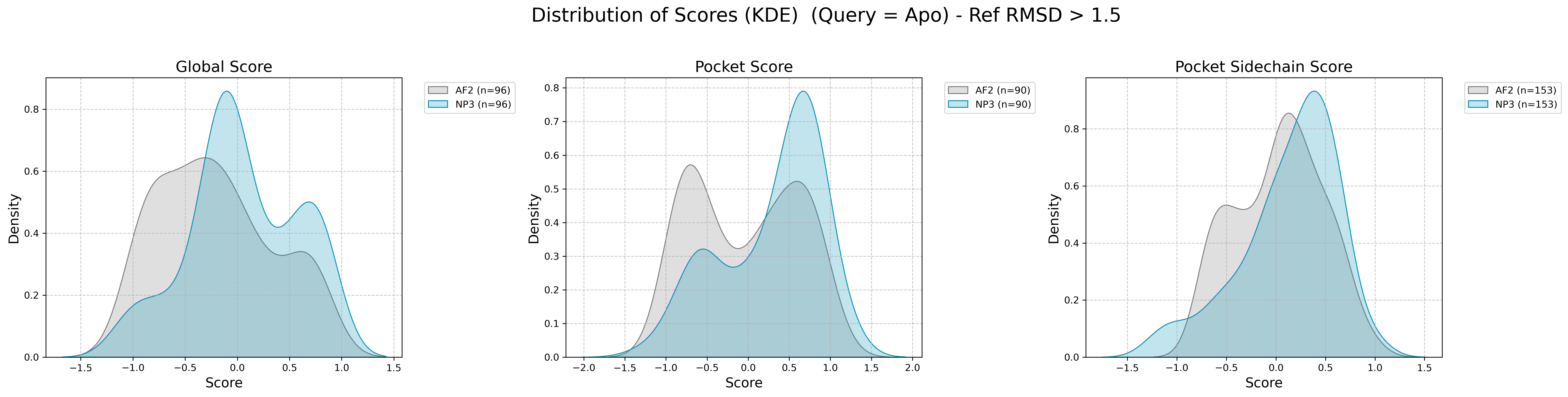}
    \caption{Kernel density estimation plots depicting the score probability densities for NP3 and AF2 models where query conformation = apo and reference RMSD > 1.5 Å.}
    \label{fig:enter-label}
\end{figure}

\begin{figure}[h]
    \centering
    \includegraphics[width=0.95\linewidth]{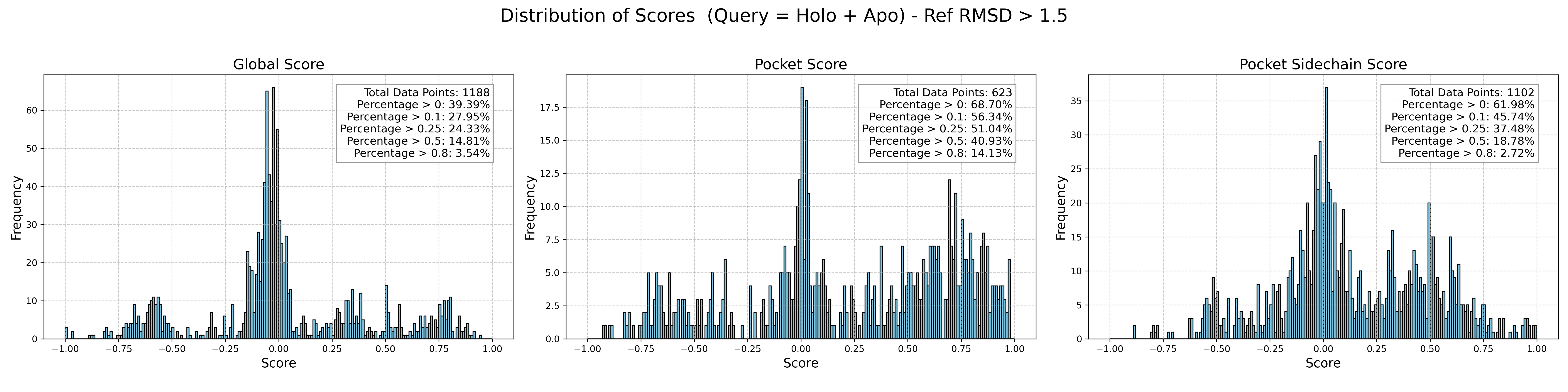}
    \caption{Distribution of ConfBench score values for NP3 with query conformation = apo or holo and a reference RMSD > 1.5 Å. The histograms display the frequency of scores across all metrics (global, pocket, and pocket-sidechain) within the range of -1 to 1.}
    \label{fig:enter-label}
\end{figure}

\clearpage

\printbibliography

\end{refsection}

\end{document}